\documentclass{article}

\usepackage{microtype}
\usepackage{graphicx}
\usepackage{subfigure}
\usepackage{booktabs} 

\usepackage{hyperref}

\usepackage[accepted]{icml2021}

\icmltitlerunning{Structured Convolutional Kernel Networks for Airline Crew Scheduling}

\usepackage[utf8]{inputenc} 
\usepackage[T1]{fontenc}    
\usepackage{url}            
\usepackage{booktabs}       
\usepackage{amsfonts}       
\usepackage{nicefrac}       
\usepackage{microtype}      

\hypersetup{ %
	pdftitle={Structured Convolutional Kernel Networks for Airline Crew Scheduling},
	pdfkeywords={},
	pdfborder=0 0 0,
	pdfpagemode=UseNone,
	colorlinks=true,
	linkcolor=blue, 
	citecolor=blue, 
	filecolor=blue, 
	urlcolor=blue, 
	pdfview=FitH,
	pdfauthor={Yassine Yaakoubi, François Soumis, Simon Lacoste-Julien},
}

\usepackage{array}
\newcolumntype{L}[1]{>{\raggedright\let\newline\\\arraybackslash\hspace{0pt}}m{#1}}
\newcolumntype{C}[1]{>{\centering\let\newline\\\arraybackslash\hspace{0pt}}m{#1}}
\newcolumntype{R}[1]{>{\raggedleft\let\newline\\\arraybackslash\hspace{0pt}}m{#1}}
\usepackage{graphicx}
\usepackage[colorinlistoftodos]{todonotes}
\usepackage{amssymb}
\newtheorem{Definition}{Definition}
\newtheorem{Theorem}{Theorem}
\usepackage{amsmath}
\DeclareMathOperator*{\argmin}{arg\,min}

\def\PP{{\mathcal P}}

\def\Real{{\mathbb R}}
\def\x{{\mathbf x}}

\def\z{{\mathbf z}}
\def\Z{{\mathbf Z}}

\def\defin{:=}

\newcommand\Fcal{\mathcal F}
\newcommand\eg{\emph{e.g.}}

\def\eg{{\it e.g.}}

\def\kmone{k\text{--}1}

\usepackage{tikz}
\usetikzlibrary{positioning,decorations.pathreplacing}
\DeclareMathOperator*{\argmax}{arg\,max}

\usepackage{url}      
\usepackage{booktabs} 
\usepackage{amsfonts} 
\usepackage{nicefrac} 
\usepackage{microtype} 
\usepackage{algorithm}
\usepackage{algorithmic}
\usepackage{dsfont}

\usepackage{wrapfig}
\usepackage{lipsum}

\begin{document}

\twocolumn[
\icmltitle{Structured Convolutional Kernel Networks for Airline Crew Scheduling}



\icmlsetsymbol{equal}{*}

\begin{icmlauthorlist}
\icmlauthor{Yassine Yaakoubi}{poly,mila}
\icmlauthor{François Soumis}{poly}
\icmlauthor{Simon Lacoste-Julien \texorpdfstring{$^\dagger$}{}}{mila,udem}
\end{icmlauthorlist}

\icmlaffiliation{poly}{GERAD, Polytechnique Montr\'eal, Canada}
\icmlaffiliation{mila}{Mila, Canada}
\icmlaffiliation{udem}{Department of Computer Science and Operations Research, Universit\'e de Montr\'eal, Canada$\,\dagger$ Canada CIFAR AI Chair}

\icmlcorrespondingauthor{Yassine Yaakoubi}{YassineYaakoubi@outlook.com}
\icmlkeywords{Machine Learning, ICML}

\vskip 0.3in
]



\printAffiliationsAndNotice{}  

\begin{abstract}
Motivated by the needs from an airline crew scheduling application, we introduce structured convolutional kernel networks (Struct-CKN), which combine CKNs from~\citet{Mairal2014} in a structured prediction framework that supports \emph{constraints} on the outputs. CKNs are a particular kind of convolutional neural networks that approximate a kernel feature map on training data, thus combining properties of deep learning with the non-parametric flexibility of kernel methods. Extending CKNs to structured outputs allows us to obtain useful initial solutions on a flight-connection dataset that can be further refined by an airline crew scheduling solver. More specifically, we use a flight-based network modeled as a general conditional random field capable of incorporating local constraints in the learning process. Our experiments demonstrate that this approach yields significant improvements for the large-scale crew pairing problem (50,000 flights per month) over standard approaches, reducing the solution cost by 17\% (a gain of millions of dollars) and the cost of global constraints by 97\%.
\end{abstract}

\section{Introduction}
\label{sec:introduction}

Since crew costs are the second-highest spending source for air passenger carriers, crew scheduling is of crucial importance for airlines. The crew pairing problem (CPP) searches for a minimum-cost set of anonymous feasible pairings (rotations) from the scheduled flights, such that all flights are covered exactly once, and all airline regulations and collective agreements are respected. The complexity of this problem lies in the large number of possible pairings, as the selection of pairings at minimal cost—a large integer programming problem—cannot be performed using standard solvers.
Seeking to obtain an efficient algorithm for large-scale monthly CPPs (up to 50,000 flights) and building on the column generation-based solver by~\citet{desaulniers2020}, \citet{Yaakoubi2020} proposed Commercial-GENCOL-DCA, an improved solver starting with an aggregation, in clusters, of flights. The initial aggregation partition permits replacing all flight-covering constraints of flights in a cluster by a single constraint, thus allowing the solver to cope with larger instances.
Initial clusters can either be extracted from the initial solution~\cite{desaulniers2020} or given separately, as in~\citet{Yaakoubi2020}, where authors used convolutional neural networks (CNN) to solve the flight-connection problem (a supervised multi-class classification problem). The objective of this problem is to predict the next flight that a crew follows in its schedule given the previous flight.
They used CNN to harness the spatial locality (localized spatial features) and used a similarity-based input, where neighboring factors have similar features. By passing initial clusters of flights to the CPP solver, the reported reduction of solution cost averages between 6.8\% and 8.52\%, mainly due to the reduction in the cost of global constraints between 69.79\% and 78.11\%.
The cost of global constraints refers to the penalties incurred when the workload is not fairly distributed among the bases in proportion to the available personnel at each base.

However, a major weakness of their approach is that they can only produce initial clusters and not an initial solution, since they use a greedy predictor making one prediction at a time (predicting sequentially the next flight given only the previous flight).
This prevents the predictor from incorporating constraints on the output and the produced solutions cannot be used as initial solutions for the solver as they are not sufficiently close to being feasible. A pairing is deemed feasible if it satisfies safety rules and collective agreement rules~\citep{kasirzadeh2017}; examples include minimum connection time between two flights, minimum rest time, and maximum number of duties in a pairing.
By providing an initial solution, we not only accelerate the optimization process and calculate the feasibility of proposed pairings, but we also propose clusters similar to the initial solution, thus reducing the degree of incompatibility between current solution and proposed pairings~\citep{Elhallaoui2008_multi}.

In this paper, we address this lack of constraint modeling, while still enabling the use of a convolutional architecture. For this purpose, we investigate the convolutional kernel network (CKN), an approximation scheme similar to CNN proposed by~\citet{Mairal2014}. To bypass the major limitations in~\citet{Yaakoubi2020}, we incorporate local constraints on the outputs (imposing that each flight has to be preceded by at most one flight). We harness the spatiotemporal structure of the CPP problem by combining kernel methods and structured prediction. The outputs start the optimizer and solve a large-scale CPP, where small savings of a mere 1\% translate into an increase of annual revenue for a large airline by dozens of millions of dollars. Note that, to the best of our knowledge, we are not aware of any ML approach that can directly solve the CPP (which has complex airline-dependent costs and constraints that are not necessarily available to the ML system at train time).

We thus consider instead to use the ML system to propose good initial clusters and an initial solution for the CPP solver. The results of training on the flight-connection dataset~\citep{Yaakoubi2019}, a flight-based network structure modeled as a general conditional random field (CRF) graph, demonstrate that the proposed predictor is more suitable than other methods. Specifically, it is more stable than CNN-based predictors and extensive tuning is not required, in that no Bayesian optimization (to find a suitable configuration) is needed, as we observe in our experiments. This is crucial to integrate ML into a solver for the CPP or any real-world scheduling problem. Note that unlike recurrent neural networks (RNNs) or neural networks by \cite{Yaakoubi2020} which cannot produce initial solutions that are sufficiently close to being feasible, the proposed predictor incorporates local constraints in the learning process. Furthermore, note that while previous studies focused on using ML (either through imitation learning or reinforcement learning) to solve small-scale CO problems such as vehicle routing ($\leq$ 100 customers) and airline crew scheduling ( $\leq$ 714 flights) problems, we use the proposed predictor to warm-start a monthly CPP solver (up to 50,000 flights). For an extensive literature review on using machine learning for combinatorial optimization, see \citet{bengio2020}.

\paragraph{Contributions.}
Bridging the gap between kernel methods and neural networks, we propose the structured convolutional kernel network (Struct-CKN)\footnote{The code is available at the following link:~\url{https://github.com/Yaakoubi/Struct-CKN}}.
We first sanity check the approach on the OCR dataset~\citep{Taskar2004} yielding a test accuracy comparable to state of the art.
Then, to warm-start an airline crew scheduling solver, we apply the proposed method on a flight-connection dataset, modeled as a general CRF capable of incorporating local constraints in the learning process. We show that the constructed solution outperforms other approaches in terms of test error and feasibility, an important metric to initialize the solver. The predicted solution is fed to the solver as an initial solution and initial clusters, to solve a large-scale CPP (50,000 flights). Our experiments demonstrate that this approach yields significant improvements, reducing the solution cost by 17\% (a gain of millions of dollars) and the cost of global constraints by 97\%, compared to baselines.

\paragraph{Outline.}
The remainder of this paper is structured as follows. Section~\ref{sec:related_work} describes related methods. Section~\ref{sec:ckn} presents CKNs. CRFs are outlined in Section~\ref{sec:structure_crfs}. Section~\ref{sec:sckn_framework} presents Struct-CKN. Section~\ref{sec:experiments} reports Computational results on OCR dataset, flight-connection dataset, and CPP.

\section{Related Work}\label{sec:related_work}

Upon a succinct review of previous work to compare available approaches in the literature to Struct-CKN, we argue for the use of the latter on the flight-connection dataset to solve CPPs.

Combining networks and energy-based models is a well-known approach since the 1990s. For instance, \citet{bottou2012} introduced graph transformer networks trained end-to-end using weighted acyclic directed graphs to represent a sequence of digits in handwritten character recognition. Furthermore, inspired by Q-learning, \citet{gygli2017} used an oracle value function as the objective for energy-based deep networks, and~\citet{belanger2016} introduced structured prediction energy networks (SPENs) to address the inductive bias
and to learn discriminative features of the structured output automatically. By assigning a score to an entire prediction, SPENs take into consideration high-order interactions between predictors using minimal structural assumptions. Nevertheless, due to the non-convexity, optimizing remains challenging, which may cause the learning model to get stuck in local optima. Another approach is to move step by step and predict one output variable at a time by applying the information gathered from previous steps. The linking between the steps is learned using a predefined order of input variable where the conditional is modeled with RNNs~\citep{zheng2015}. Although this method has achieved impressive results in machine translation~\citep{leblond2017}, its success ultimately depends on the neural network's ability to model the conditional distribution and it is often sensitive to the order in which input data is processed, particularly in large-size graphs, as in CPPs (50,000~nodes).

In contrast to these approaches, instead of using continuous relaxation of output space variables~\citep{belanger2016}, Struct-CKN uses supervised end-to-end learning of CKNs and CRF-based models. Accordingly, any of the existing inference mechanisms—from belief propagation to LP relaxations—can be applied. This allows us to naturally handle general problems that go beyond multi-label classification, and to apply standard structured loss functions (instead of extending them to continuous variables, as in the case of SPENs). More importantly, Struct-CKN allows us to apply our method to a large-scale CRF graph containing up to 50,000 nodes. Furthermore, in contrast to methods in the literature (e.g., CNN-CRF~\citep{CRFCNN}, CRF-RNN~\citep{CRFRNN}, and deep structured models~\citep{chen2015}), it has far fewer parameters, thus bypassing the need for extensive tuning.

Finally, note that in recent papers, convolutional graph neural networks (ConvGNNs)~\citep{Kipf} are used either (1) to warm-start a solver (trained under the imitation learning framework)~\cite{owerko2020}, (2) to solve the optimization problem end-to-end~\citep{khalil2017}, or (3) to guide an optimization process (variable selection in branch-and-bound~\citet{gasse2019}), and ConvGNNs might appear to be a good candidate-solution for CPPs when coupled with a CRF layer to impose constraints on the output. However, in the case of graphs with up to 50,000 nodes, the number of parameters used by ConvGNN and the computational limitations prevents us from considering it. In fact, we are not aware of any prior work where ConvGNNs were used at this scale. Furthermore, the implicit motivation for our proposed approach is an end-to-end solution method that can (1) harness the predictive capabilities of the ML predictor and the decomposition capacity of the solver~\citep{Yaakoubi2020}, and (2) can be used on a standard machine with no specific resource requirements, to replace existing solvers in the industry.
Future research will look into the possibility of integrating a distributed version of ConvGNNs into the proposed framework.

\section{Convolutional Kernel Networks}
\label{sec:ckn}

CKN is a particular type of CNN that differs from the latter in the cost function to be optimized to learn filters and in the choice of non-linearities.
We review CKNs, with the same notation as in~\citet{Mairal2016,Bietti2019}. For further detail, see Appendix~\ref{appendix:ckn}.

\subsection[Unsupervised Convolutional Kernel Networks]{Unsupervised Convolutional Kernel Networks~\cite{Mairal2014}}
\label{sec:unsup_ckn}

We consider an image~$I_0: \Omega_0 \to \Real^{p_0}$, where $p_0$ is the number of channels, \eg, $p_0=3$ for RGB, and $\Omega_0 \subset [0,1]^2$ is a discrete set of pixel locations.
Given two image patches $x$, $x'$ of size $e_0 \times e_0$, represented as vectors in $\Real^{p_0 e_0^2}$, we define a kernel
$K_1(x,x') = \|\ x \|  \|\ x' \| \cdot \kappa_1( \langle \tfrac{x}{\|\ x \|} , \tfrac{x'}{\|\ x' \|} \rangle )$ if $x$, $x'$ $\neq$ $0$ and $0$ otherwise, where $\|.\|$ and $\langle , \rangle$ denote the Euclidian norm and inner-product, respectively, and $\kappa_1(\langle \cdot,\cdot\rangle)$ is a dot-product kernel on the sphere.
We have implicitly defined the reproducing kernel Hibert space (RKHS) $\mathcal{H}_1$ associated to $K_1$ and a mapping $\varphi_1:\Real^{p_0 e_0^2} \to \mathcal{H}_1$.

First, we build a database of $n$ patches $\x_1, \ldots,\x_n$ randomly extracted from various images and normalized to have unit $\ell_2$-norm.
Then, we perform a spherical $K$-means algorithm~\citep{buchta2012}, acting as a Nystr\"om approximation, to obtain $p_1$ centroids $\z_1,\ldots,\z_{p_1}$ with unit $\ell_2$-norm.
Given a patch~$\x$ of $I_0$, the projection of~$\varphi_1(\x)$ onto~$\Fcal_1 \defin \text{Span}(\varphi_1(\z_1),\ldots, \varphi_1(\z_{p_1}))$ admits a natural parametrization given in~\eqref{eq:param} where $\Z=[\z_1,\ldots,\z_{p_1}]$, and $\kappa_1$ is applied pointwise to its arguments.

\vspace{-0.65cm}
\begin{equation}
\begin{split}
\Gamma_1(\x) & \defin  \| \x \| \kappa_1(\Z^{\top} \Z)^{-1/2} \kappa_1\left(\Z^\top\frac{\x}{\|\ x \|}\right) \\
           &~\text{if}~~\x \neq 0~~\text{and}~~0~~\text{o.w.}
\end{split}
\label{eq:param}
\end{equation}

Consider all overlapping patches of~$I_0$. We set
$M_1(z) = \Gamma_1(\x_z), \quad z \in \Omega_0$ where $\x_z$ is the patch from $I_0$ centered at pixel location $z$. The spatial map $M_1: \Omega_0 \to \Real^{p_1}$ thus computes the quantities $\Z^\top \x$ for all patches $\x$ of image~$I$ (spatial convolution after mirroring the filters $\z_j$), then applies the pointwise non-linear function~$\kappa_1$.

The previous steps transform the image $I_0: \Omega_0 \to \Real^{p_0}$ into a map $M_1: \Omega_0 \to \Real^{p_1}$. Then, the CKNs involve a pooling step to gain invariance to small shifts, leading to another finite-dimensional map $I_1:
\Omega_1 \to \Real^{p_1}$ with a smaller resolution: $I_1(z) =  \sum_{z' \in \Omega_0} M_1(z') e^{-\beta_1 \|z'-z\|_2^2}, \quad z \in \Omega_1$, where $\beta_1$ is a subsampling factor. We build a multilayer image representation by stacking and composing kernels. 
Similarly to the first CKN layer transforming $I_0: \Omega_0 \to \Real^{p_0}$ to the map $I_1: \Omega_1 \to \Real^{p_1}$, we apply the same procedure to obtain $I_2: \Omega_2 \to \Real^{p_2}$, where $p_2$ is the number of centroids in the second layer, then $I_3: \Omega_3 \to \Real^{p_3}$, etc.

\subsection[Supervised Convolutional Kernel Networks]{Supervised Convolutional Kernel Networks~\citep{Mairal2016}}
\label{sec:sup_ckn}

Let $I_0^1$, $I_0^2$, \ldots, $I_0^n$ be the training images with respective labels $y_1$, \ldots, $y_n$ in \{-1 ; +1\} for binary classification. We also have $L$: $\mathbb{R} \times \mathbb{R} \to \mathbb{R} $, a convex smooth loss function. 
Given a positive definite kernel $K$ on images, the classical empirical risk minimization formulation consists of finding a prediction function in the RKHS $\mathcal{H}$ associated to $K$ by minimizing the objective
$\min_{f \in \mathcal{H}}  \frac{1}{n} \sum_{i=1}^{n} L(y_i , f(I_0^i)) + \frac{\lambda}{2} \left \lVert f \right \rVert_\mathcal{H}^2$, where the parameter $\lambda$ controls the smoothness of the prediction function $f$ with respect to the geometry induced by the kernel, hence regularizing and reducing overfitting.

After training a CKN with $k$ layers, such a positive definite kernel $K_\mathcal{Z}$ may be defined as in~\eqref{eq_pdk} where $I_k$, $I_k'$ are the $k$-th finite-dimensional feature maps of $I_0$ and $I_0'$, respectively, and $f_k$, $f_k'$ are the corresponding maps in $\Omega_k \to \mathcal{H}_k $, which have been defined in Section~\ref{sec:unsup_ckn}.
 
\begin{equation}
K_\mathcal{Z} (I_0 , I_0') = \sum_{z \in \Omega_k}  \langle f_k(z), f_k'(z) \rangle_{\mathcal{H}_k} = \sum_{z \in \Omega_k} \langle I_k(z), I_k'(z) \rangle
\label{eq_pdk}
\end{equation}

The kernel $K_\mathcal{Z}$ is also indexed by $\mathcal{Z}$, representing network parameters (subspaces $\mathcal{F}_1$, \ldots, $\mathcal{F}_k$, or equivalently the set of filters $Z_1$, \ldots, $Z_k$). Then, the formulation becomes as in~\eqref{eq7} where $\left \lVert \cdot \right \rVert_F$ is the Frobenius norm extending the Euclidean norm to matrices and, with an abuse of notation, the maps $I_k^i$ are seen as matrices in $\mathbb{R}^{p_k \times \lvert \Omega_k \lvert }$. Then, the supervised CKN formulation consists of jointly minimizing~\eqref{eq7} w.r.t. $W$ in $\mathbb{R}^{p_k \times \lvert \Omega_k \lvert }$ and with respect to the set of filters $Z_1$, \ldots, $Z_k$, whose columns are constrained to be on the Euclidean sphere.

\begin{equation}
\min_{W \in \mathbb{R}^{p_k \times \lvert \Omega_k \lvert } }  \frac{1}{n}  \sum_{i=1}^{n} L(y_i , \langle W , I_k^i \rangle  ) + \frac{\lambda}{2} \left \lVert W \right \rVert_F^2
\label{eq7}
\end{equation}


\section{Graph-Based Learning} 
\label{sec:structure_crfs}

In structured prediction, models are typically estimated with surrogate structured loss minimization, such as with structured SVM (SSVM) or CRFs. We used CRFs for the structured prediction that we briefly review below. We also tested SSVM integration instead of CRFs, but it yielded slightly worse results, see Appendices~\ref{appendix:struct_prediction} and~\ref{appendix:add_sckn_results} for details and experimental results using SSVM.

A CRF models the conditional probability of a structured output $y \in \mathcal{Y}$ given an input $x \in \mathcal{X}$ where the probability to observe $y$ when $x$ is observed is $ p(y | x; w) \propto \exp(\langle w, F(x,y)\rangle) $, $F$ is the feature mapping, and $w$ is the vector of weights (to be learned). The CRF predictor of $y$ when $x$ is observed is: $ h_w(x)=\argmax_{y \in \mathcal{Y}} \langle w, F(x,y) \rangle $.
The CRF primal problem formulation is shown in~\eqref{crf_primal}, where $\mathcal{L}^{CRF}$ denotes the negative log likelihood loss.

\begin{equation}
\min_{\omega}\lambda \lVert \omega \rVert^2+\frac{1}{n}\sum_{i=1}^{N}\mathcal{L}^{CRF}(x_i, y_i; \omega)
\label{crf_primal}
\end{equation}

To optimize CRFs, we use stochastic dual coordinate ascent (SDCA)~\citep{shalev2013b,shalev2014} as~\citet{Priol2018} showed it yielded state-of-the-art results for CRFs.
Although one can also use the stochastic average gradient (SAG) algorithm~\citep{schmidt2015} or the online exponentiated gradient (OEG) algorithm, an advantage of SDCA over OEG (and SAG) is that it enables performing an ``exact'' line search with only one call to the marginalization oracle.
We now rewrite~\eqref{crf_primal} using the notation for the SDCA setup for multi-class classification~\citep{shalev2014}.
Denote $M_i = | \mathcal{Y}_i |$ the number of labelings for sequence $i$.
Denote $A_i$ the matrix whose columns are the corrected features $\{\psi_i(y)\defin F(x_i,y_i)-F(x_i,y)\}_{y \in \mathcal{Y}_i}$. Denote also $\Phi_i(s) \defin \log( \sum_{y \in \mathcal{Y}_i} \exp(s_y))$ the log-partition function for the scores $s \in \mathbb{R}^{M_i}$. Since the negative log-likelihood can be written as $-\log(p(y_i|x_i;w)) = \Phi_i(-A_i^\top w)$~\citep{Murphy2012}, the primal objective function to minimize over $w \in R^d$ becomes
$
P(w) \defin \frac{\lambda}{2} \left \lVert w \right \rVert_2^2 + \frac{1}{n} \sum_{i=1}^{n} \Phi_i(-A_i^\top w).
$
This minimization problem has an equivalent Fenchel convex dual problem.
Denote $\Delta_M$ the probability simplex over $M$ elements.
Denote $\alpha_i \in \Delta_{M_i}$ the set of dual variables for a given $x_i$,
we define the conjugate weight function $\hat{w}$ as follows: $\hat{w}(\alpha) = \tfrac{1}{\lambda n} \sum_i A_i \alpha_i = \tfrac{1}{\lambda n} \sum_{i=1}^{n} \mathbb{E}_{y \sim \alpha_i} (\psi(y))$.
We can show that $\hat{w}(\alpha^*) = w^*$ where $w^*$ and $\alpha^*$ are respectively the optimal primal parameters and the optimal dual parameters. As such, we can also define the primal sub-optimality as: $P(w) - P(w^*)$.

\citet{Priol2018} adapted SDCA to CRF by considering marginal probabilities over cliques of the graphical model. Because the dual variable $\alpha_i$ is exponentially large in input size $x_i$, $\alpha$ is replaced by $\mu$ = ($\mu_1$, \ldots, $\mu_n$), where $\mu_i \in \Pi_C^{\Delta_C}$ is the concatenation of all the clique marginal vectors for sample~$i$. Given its state-of-the-art performance, we decided to use it, although we will compare it with other optimizers in Section~\ref{sec:experiments} (see Appendix \ref{appendix:crf} for further details).

\section{The Struct-CKN Framework}
\label{sec:sckn_framework}

\begin{figure*}
\centering
\input{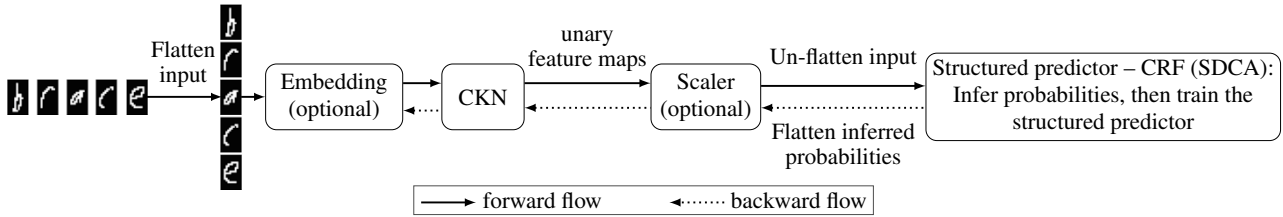}
\caption{The Architecture Diagram for the Struct-CKN Predictor}
\label{fig:flowchart}
\end{figure*}

As in Figure~\ref{fig:flowchart}, the Struct-CKN model consists of two components intended to train CRFs: (1) the CKN and (2) the structured predictor, using CRF loss, and SDCA. Upon initializing the CKN layers and the structured predictor, for each iteration, we pass the input image through the CKN multilayers. The last map of CKN is passed through to the structured predictor to infer probabilities, which are employed to train CKN weights by backpropagating using rules in~\citet{Mairal2016}.

\begin{algorithm}
	\caption{Training the Struct-CKN Model}\label{alg:sckn}
	\begin{algorithmic}[1]
	   \STATE Initialize CKN parameters in unsupervised manner as described in Sec.~\ref{sec:unsup_ckn}
	   \STATE Initialize structured predictor and CRF model as in Alg.~\ref{alg-sdca} (see Appendix \ref{appendix:crf}, steps 1-2)
	   \FOR{$t=0\dots$}
	        \STATE
	        For each input, construct an unary feature map, as described in Sec.~\ref{sec:unsup_ckn}
	        \STATE (Optional) Center and rescale these representations to have unit $\ell_2$-norm on average
	        \STATE Infer probabilities by providing image map as input to structured predictor
	        \STATE Train the structured predictor using the feature map as an input for $n_{Ep}$ epochs
	        \STATE Use the inferred probabilities to compute the gradient by using the chain rule (backpropagation) and update the CKN weights~\citep{Mairal2016}
	   \ENDFOR
	\end{algorithmic}
\end{algorithm}

To do inference for CRF models with a small number of nodes (as in Section \ref{sec:ocr}), we use max-product belief propagation, since chains can be solved exactly and efficiently. For the flight-connection dataset (as in Section \ref{sec:airline}), since CRF models contain 50,000 nodes, we use AD3 (alternating directions dual decomposition)~\citep{martins2015} for approximate maximum a posteriori (MAP) inference. It allies the modularity of dual decomposition with the effectiveness of augmented Lagrangian optimization via the alternating directions method of multipliers and has some very interesting features in comparison to other message-passing algorithms. Indeed, AD3 has been empirically shown to reach consensus faster than other algorithms~\citet{martins2015} and outperforms state-of-the-art message-passing algorithms on large-scale problems. Besides, AD3 provides a library of computationally-efficient factors that allow handling declarative constraints within an optimization problem. This is particularly interesting for the CPP use case since we add a constraint to impose that each flight is preceded by one flight at most. To use SDCA (requiring a marginalization oracle) with AD3 (used to do approximate MAP), we propose a simple approximation, using MAP label estimates. See Appendix \ref{appendix:crf} for details on integrating SDCA and AD3.

Note that an embedding layer may be used before passing the input through to the CKN layers. Indeed, for categorical variables with a large number of categories, the input matrix is sparse, making the learning process difficult, as the extracted patches have mostly null values. Furthermore, as in~\citet{chandra2017}, using deep structured predictors may require using scaling. Specifically, the last map of the ``deep'' layer needs to be rescaled before being passed to the ``structured layer''. We propose to use one of the following scalers within the Scikit-learn library: Min-Max scaler, Normalizer scaler, Standard scaler, and Robust scaler.
When using the SDCA optimizer, line search requires computing the entropy of the marginals. Since this is costly, in order to minimize the number of iterations, we used the Newton-Raphson algorithm. This requires storing the logarithm of the dual variable, which may be expensive, so a decent amount of memory should be allocated.

Finally, note that we use a batch-version of the Struct-CKN predictor on the flight-connection dataset~\citep{Yaakoubi2019}, where one batch corresponds to one CRF model (CPP instance): (1) We initialize the CKN weights (in an unsupervised manner) and the CRF model sequentially by considering each one of the six instances separately. (2) We ''flatten'' the input by considering each one of the six instances separately. (3) We center and rescale the representations for all the instances at once. (4) We pass small batches of image maps as inputs to the structured predictor (e.g., 128 image maps) to infer the probabilities and train the structured predictor. Then, we use the inferred probabilities of the batch to update the CKN weights.

\section{Experiments}
\label{sec:experiments}

In this section, we report the results of experiments using Struct-CKN.
First, we sanity-check Struct-CKN on the standard OCR dataset in Section \ref{sec:ocr}, showing that it's comparable to the state of the art. Then, in Section \ref{sec:airline}, we use the proposed predictor on the flight-connection dataset to warm-start Commercial-GENCOL-DCA.
We use Pytorch~\cite{pytorch} to declare said model and perform operations on a 40-core machine with 384\,GB of memory, and use K80 (12\,GB) GPUs. 
The CRF model is implemented using PyStruct~\cite{pystruct}, while the SDCA optimizer is implemented using SDCA4CRF. Scalers are implemented using Scikit-learn~\cite{scikit-learn}.

\subsection{OCR - Chain CRF}\label{sec:ocr}

\begin{table}
\centering
\caption{Test Error on the OCR Dataset}
\resizebox{\linewidth}{!}{ 
\begin{tabular}{lr}
\toprule
 & Test error (\%) \\
\midrule
SDCA - linear features~\citep{Priol2018} & 12.0\\
\midrule
LSTM~\citep{lstm} & 4.6 \\
CNN-CRF~\citep{CRFCNN} & 4.5 \\
SCRBM~\citep{Tran2018} & 4.0 \\
NLStruct~\citep{graber2018}& 3.6 \\
\midrule
Struct-CKN & \textbf{3.4} \\
\bottomrule
\end{tabular}
}
\label{table-test-ocr}
\end{table}

\begin{figure}
\centering
\includegraphics[width=\linewidth]{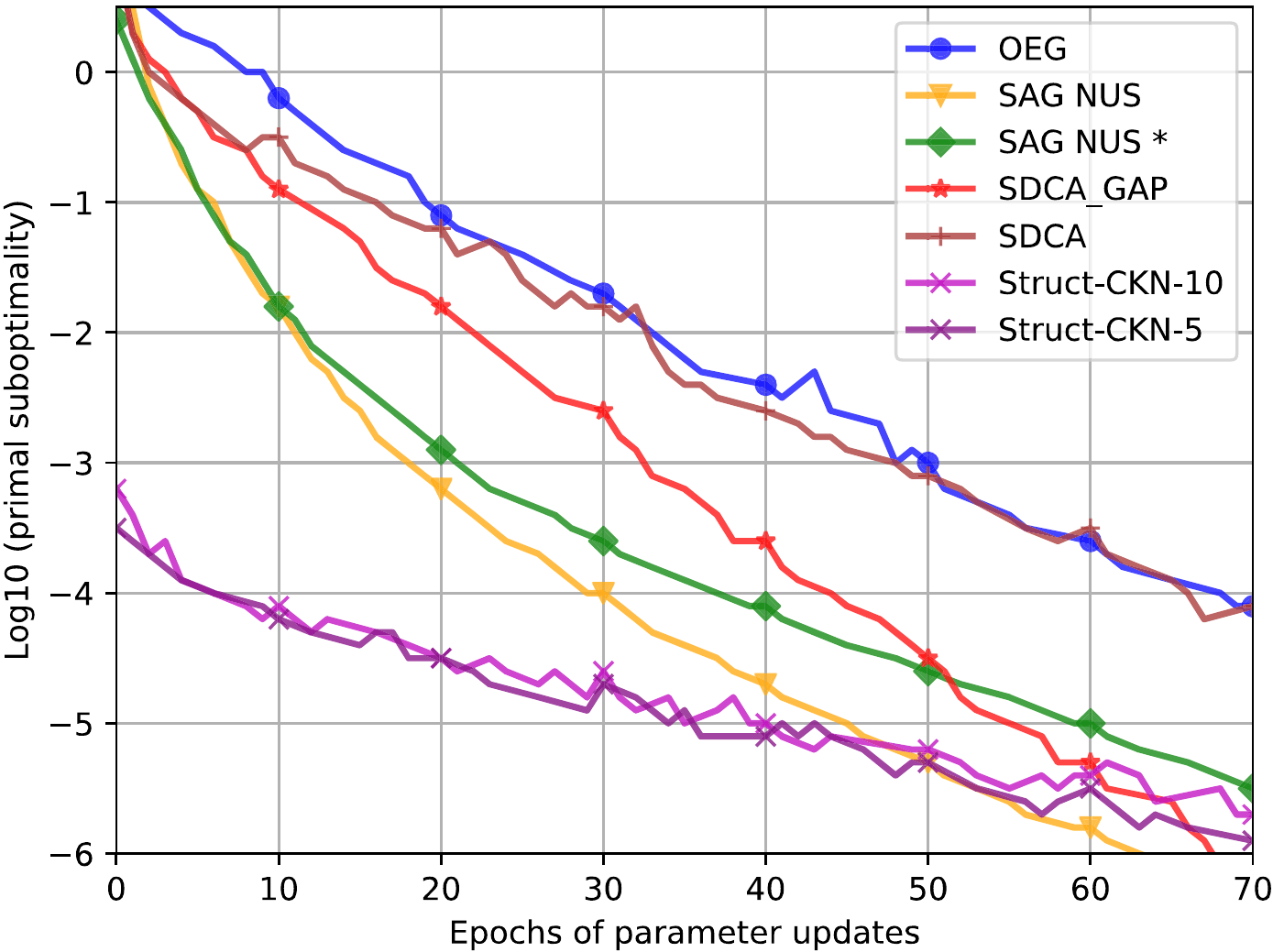}
\vspace{-2mm}
\caption{Comparison of Primal Sub-optimality}
\label{fig:subopt_comparison}
\end{figure}

Each example in the OCR dataset~\citep{Taskar2004} consists of a handwritten word pre-segmented into characters, with each character represented as a 16$\times$8 binary image. The task is to classify the image into one of the 26 characters (a$-$z). It comes with pre-specified folds; one fold is considered the test set, while the rest as the training set, as in max-margin Markov networks~\citep{Taskar2004}.
Since the CRF optimizers (SAG-NUS, SAG-NUS*~\citep{schmidt2015}, SDCA, SDCA-GAP~\citep{Priol2018}, and OEG~\citep{schmidt2015}) yield similar test errors (11.8-12\%), we only report SDCA (with linear features) in Table~\ref{table-test-ocr}. LSTM (standard two-layer)~\citep{lstm} and CNN-CRF (standard two-layer)~\citep{CRFCNN} yield comparable results (4.4-4.6\%). Sequence Classification Restricted Boltzmann Machine (SCRBM)~\citep{Tran2018} and NLStruct~\citep{graber2018} provide better results (4.0\%, and 3.6\%). However, Struct-CKN outperforms all aforementioned methods, reducing test errors to 3.40\%.

Note that some structured predictors can lower test error to $1-3\%$~\citep{perez2007}, such as SeaRNN~\citep{leblond2017}, which adapts RNN to the learning-to-search approach. However, such models approximate the cost-to-go for each token by computing the task loss for as many roll-outs as the vocabulary size at each time step, and are thus difficult to scale to real-world datasets (with long sequences or large vocabulary), such as the flight-connection dataset (see Section~\ref{sec:airline}). Figure~\ref{fig:subopt_comparison} reports primal sub-optimality w.r.t parameter updates (see Appendix~\ref{appendix:add_sckn_results}). Struct-CKN outperforms other methods for the first $50$ epochs and is comparable to other methods for subsequent epochs. We usually train predictors only for several epochs; thus, the precision of primal sub-optimality (below $10^{-5}$) is negligible.

\subsection{Airline Crew Scheduling Dataset - Graph CRF}\label{sec:airline}

This section reports results of using Struct-CKN to warm-start Commercial-GENCOL-DCA~\citep{Yaakoubi2020} and solve large-scale CPP. Section~\ref{sec:cpp} presents CPP. Section~\ref{prediction_problem_form_sec} outlines the flight-connection prediction problem and CRF models. Section~\ref{ml_results} reports results of predictions. Section~\ref{sec:feasibility} describes the optimization process and analyzes the feasibility of proposed monthly solutions. Section~\ref{sec:or_results} reports results of solving CPPs using the solver.

\subsubsection{Crew Pairing Problem}\label{sec:cpp}

The CPP aims to find a set of pairings at minimal cost for each category of the crew and each type of aircraft fleet~\citep{Desaulniers1997}. A flight sequence operated by a single crew forms a duty and a consecutive sequence of duty periods is named a pairing.
A pairing is deemed feasible if it satisfies safety rules and collective agreement rules~\citep{kasirzadeh2017}, such as:
\begin{itemize}
\item minimum connection time between two consecutive flights and minimum rest-time between two duties;
\item maximum number of flights per duty and maximum span of a duty;
\item maximum number of landings per pairing and maximum flying time in a pairing;
\item maximum number of days and maximum number of duties in a pairing.
\end{itemize}
%

In addition, CPPs use base constraints (referred to as global constraints) to distribute the workload fairly amongst the bases proportionally to the personnel available at each base. Penalties when the workload is not fairly distributed is called the cost of global constraints.
In our approach, the predictor does not consider these very complex airline-dependent constraints as well as the solution cost and the cost of global constraints, as doing so would require more data than are currently available. Whenever a flight appears in more than one pairing, we use deadheads: one crew operates the flight, while the others are transferred between two stations for repositioning.

The CPP has been traditionally modelled as a set partitioning problem, with a covering constraint for each flight and a variable for each feasible pairing~\cite{Desaulniers1997, kasirzadeh2017}.
Formally, we consider $F$ to be a set of flights that must be operated during a given period and $\Omega$ to be the set of all feasible pairings that can be used to cover these flights.
It is computationally infeasible to list all pairings in $\Omega$ when solving CPPs with more than hundreds of flights. Therefore, it is not tractable to do so in this context (CPPs with 50,000 flights).
For each pairing $p \in \Omega$, let $c_p$ be its cost and $a_{fp}$, $f \in F$, be a constant equal to $1$ if it contains leg $f$ and $0$ otherwise.
Moreover, let $x_p$ be a binary variable that takes value $1$ if pairing $p$ is selected, and $0$ otherwise.
Using a set-partitioning formulation, the CPP can be modelled as follows:

\begin{align}
& \underset{x}{\text{minimize}}
& \sum_{p \in \Omega}{c_{p} x_p} \label{Eq:CPP1} \\
& \text{subject to}
& \sum_{p \in \Omega}{a_{fp} x_p} = 1
& & \forall f \in F   \label{Eq:CPP2} \\
&
& x_p \in \{ 0 , 1 \}
& & \forall p \in \Omega \label{Eq:CPP3}
\end{align}

The objective function (\ref{Eq:CPP1}) minimizes the total pairing costs. Constraints (\ref{Eq:CPP2}) ensure each leg is covered exactly once, and constraints (\ref{Eq:CPP3}) enforce binary requirements on the pairing variables.
The methodology to solve CPPs depends on the size of the airline's network, rules, collective agreements, and cost structure~\citep{Birge2006}. Since the 1990s, the most prevalent method has been column generation inserted in branch-\&-bound~\citep{Desaulniers1997}. This algorithm was combined with multiple methods in~\citet{desaulniers2020} to solve large-scale CPPs. \citet{Yaakoubi2020} proposed Commercial-GENCOL-DCA with a dynamic control strategy and used CNNs to develop initial monthly crew pairings. Because the constructed solution contained too many infeasible pairings, it was passed to the solver only as initial clusters and a generic standard initial solution was used.
In this work, we first use Struct-CKN to construct initial monthly crew pairings passed to the solver both as initial clusters and an initial solution. 
First, we use the GENCOL solver (used to assign crews to undercovered flights),\footnote{\url{http://www.ad-opt.com/optimization/why-optimization/column-generation/}} then we run Commercial-GENCOL-DCA.
Because solvers can only handle a few thousand flights, the windowing approach is used: the month is divided into multiple windows, where each window is solved (sequentially) while flights in pairings from previous windows are frozen.
GENCOL requires using two-day windows and one-day overlap period, while Commercial-GENCOL-DCA permits to use one-week windows and two-day overlap period.
The latter starts with an aggregation, in clusters, of flights. The initial aggregation partition permits replacing all flight-covering constraints of the flights in a cluster by a single constraint, allowing to cope with larger instances.

\subsubsection{Prediction Problem Formulation}\label{prediction_problem_form_sec}

We aim to provide the CPP solver with an initial solution and initial clusters. Using the flight-connection dataset built in~\citet{Yaakoubi2019}, the flight-connection prediction problem is a multi-class classification problem, formulated as follows: ``Given the information about an incoming flight in a specific connecting city, choose among all possible departing flights from this city the one that the crew should follow'' (see Appendix~\ref{appendix:cpp}).
Thus, each input contains information on the previous flight performed by the crew as well as information on all possible next flights (up to 20 candidates), sorted by departure time, and the output (class) is the rank of the flight performed by the crew in past solutions. 
In~\citet{Yaakoubi2019}, the authors propose a similarity-based input where neighboring factors have similar features, allowing the use of CNNs.
We extend their approach with our Struct-CKN model, thus starting the CPP solver with an initial solution, and not only the initial clusters.

The training set in the flight-connection dataset consists of six monthly crew pairing solutions (50,000 flights per month) and the test set is a benchmark that airlines use to decide on the commercial solver to use. Each flight is characterized by the cities of origin and destination, the aircraft type, the flight duration, and the departure and arrival time. For each incoming flight, the embedded representation of the candidate next flights is concatenated to construct a similarity-based input, where neighboring factors have similar features. The intuition is that each next flight is considered a different time step, enabling the use of convolutional architecture across time. \citet{Yaakoubi2019} compare multiple predictors on the flight-connection dataset and empirically confirm this intuition (see Appendix \ref{appendix:cpp}).

The dataset is used to define a pairwise CRF on a general graph where each example consists of a flight-based network structure with nodes corresponding to flights and \emph{arcs representing the feasibility of two flights being successive}. Each flight corresponds to a node connected to nodes that are possible successors/predecessors; the true label is the rank of the next flight in the set of sorted possible successors. We impose local constraints to the output by imposing that each flight has to be preceded by at most one flight (using a XOR constraint, which contributes $-\infty$ to the potential).
Unlike for the OCR dataset where “max-product” is used for belief propagation, in this section, we use AD3 (alternating directions dual decomposition)~\citep{martins2015} for approximate maximum a posteriori (MAP) inference.

\subsubsection[Results on the Flight-Connection Dataset]{Results on the Flight-connection Dataset}\label{ml_results}

As Struct-CKN has fewer hyperparameters than CNNs, we observed that Struct-CKN is more stable than CNNs in our experiments, in that it does not requires Bayesian optimization to find a good architecture, thus justifying our use of CKN (see Appendix~\ref{appendix:add_sckn_results}). Table~\ref{table:test_acc_fcd} reports the number of parameters and the test error on the flight-connection dataset using (1) CNNs and Bayesian optimization to search for the best configuration of hyperparameters~\citep{Yaakoubi2020}; (2) standard CNN-CRF (with non-exhaustive hyperparameter tuning)~\citep{CRFCNN}; and (3) Struct-CKN. Struct-CKN outperforms both CNN and CNN-CRF while having far fewer parameters (97\% fewer parameters). While Bayesian optimization can be used to fine-tune hyperparameters, this is not feasible in a real-case usage scenario, as practitioners cannot perform it each time new data become available and the need for extensive fine-tuning makes a predictor impossible to integrate into any scheduling solver.

\begin{table}
\centering
\caption{Test Error on the Flight-connection Dataset~\citep{Yaakoubi2019}}
\label{table:test_acc_fcd}
\resizebox{\columnwidth}{!}{
\begin{tabular}{lcr}
\toprule
& Test error (\%) & \# parameters \\
\midrule
CNN~\citep{Yaakoubi2020} & 0.32 & 459 542 \\
\hline
CNN-CRF~\citep{CRFCNN} & 0.38 & 547 542 \\
\hline
Struct-CKN & \textbf{0.28} & \textbf{15 200} \\
\bottomrule
\end{tabular}
}
\end{table}

\subsubsection{Construction and Feasibility of a Monthly Solution}\label{sec:feasibility}

\begin{figure}
\centering
\begin{tikzpicture}[scale=0.65, every node/.style={scale=0.7}]

    \node[rectangle, minimum width=10cm, minimum height=0.5cm, draw, line width=0.8pt] (R1) at (0,0) {break all illegal pairings};
    \node[rectangle, minimum width=10cm, minimum height=0.5cm, align=center, draw, line width=0.8pt, anchor=north] (R2) at ([yshift=-1cm]R1.south) {Generate deadheads and solve with OR solver \\ (GENCOL then Commercial-GENCOL-DCA)};

    \draw[line width=1pt] ([xshift=1cm]R1.north west)--([yshift=0.5cm,xshift=1cm]R1.north west)node[above,align=center]{Standard -\\ initial};
    \draw[line width=1pt] ([xshift=1cm]R1.south west)--([xshift=1cm]R2.north west)node[midway,fill=white]{Standard-feasible};
    \draw[line width=1pt, -latex] ([xshift=1cm]R2.south west)--([xshift=1cm,yshift=-1.5cm]R2.south west)node[midway,fill=white]{Baseline};

    \draw[line width=1pt] ([xshift=-7.1cm]R1.north east)--([yshift=0.5cm,xshift=-7.1cm]R1.north east)node[above,align=center]{CNN -\\ initial};
    \draw[line width=1pt] ([xshift=-5.3cm]R1.north east)--([yshift=0.5cm,xshift=-5.3cm]R1.north east)node[above,align=center]{CNN};    
    \draw[line width=1pt] ([xshift=-3.2cm]R1.north east)--([yshift=0.5cm,xshift=-3.2cm]R1.north east)node[above,align=center]{Struct-CKN -\\ initial};
    \draw[line width=1pt] ([xshift=-7.1cm]R1.south east)--([xshift=-7.1cm]R2.north east)node[midway,fill=white]{CNN-feasible};
    \draw[line width=1pt, -latex] ([xshift=-7.1cm]R2.south east)--([xshift=-7.1cm,yshift=-1.5cm]R2.south east)node[midway,fill=white]{CNN};
    \draw[line width=1pt] ([xshift=-3.2cm]R1.south east)--([xshift=-3.2cm]R2.north east)node[midway,fill=white]{Struct-CKN-feasible};
    \draw[line width=1pt, -latex] ([xshift=-3.2cm]R2.south east)--([xshift=-3.2cm,yshift=-1.5cm]R2.south east)node[midway,fill=white]{Struct-CKN};
    
    \draw[line width=1pt] ([xshift=-0.7cm]R1.north east)--([yshift=0.5cm,xshift=-0.7cm]R1.north east)node[above]{Struct-CKN};
    \draw[line width=1pt] ([xshift=-0.7cm]R1.south east)--([xshift=-0.7cm]R2.north east);
    \draw[line width=1pt] ([xshift=-5.3cm]R1.south east)--([xshift=-5.3cm]R2.north east);
    \draw[line width=1pt, -latex] ([xshift=-0.7cm]R2.south east)--([xshift=-0.7cm,yshift=-1.5cm]R2.south east)node[midway,fill=white]{Struct-CKN\textbf{+}};
    \draw[line width=1pt, -latex] ([xshift=-5.3cm]R2.south east)--([xshift=-5.3cm,yshift=-1.5cm]R2.south east)node[midway,fill=white]{CNN\textbf{+}};

\end{tikzpicture}
\caption{The Optimization Process for the CPP}
\label{fig:opt_process}
\end{figure}
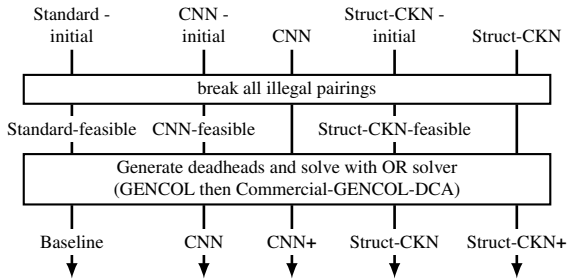

As in Figure~\ref{fig:opt_process}, we compare four approaches to construct monthly pairings. The first approach is a standard monthly solution, called \textit{``Standard - initial''}, a ``cyclic'' weekly solution (from running the optimizer on a weekly CPP) rolled to cover the whole month~\citep{desaulniers2020}. A cyclic solution is where the number of crews in each city is the same at the beginning and end of the horizon. In the second approach, CNNs predict the flight-connection probabilities. Then, using the same heuristics as in~\citet{Yaakoubi2020} (see Appendix~\ref{sec:cnn_probs_monthly_cp}), we build a monthly crew pairing called \textit{``CNN - initial''}. In the third and fourth approaches, CNN-CRF~\citep{CRFCNN} and Struct-CKN are used to build monthly crew pairings called \textit{``CNN-CRF - initial''}, and \textit{``Struct-CKN - initial''}. Then, we break all illegal pairings and freeze the legal sub-part in initial crew pairings, resulting in \textit{``Standard - feasible''}, \textit{``CNN - feasible''}, \textit{``CNN-CRF - feasible''} and \textit{``Struct-CKN - feasible''}. We generate deadheads on all flights and pass it to the solver.

Note that to reproduce the following results, we obtain the commercial dataset from~\citet{Yaakoubi2019}, which cannot be distributed because it contains too much flight-data information sensitive to airlines' operations. Since the first step of the CPP solver can solve up to several thousand flights, we are constrained to use two-day windows. Since pairings extend to over two days, it cannot ``fix'' the mispredicted and illegal pairings. Therefore, the more illegal pairings the constructed solution contains, the longer it will take the solver to find a suitable final solution. Furthermore, when the constructed solution is highly infeasible, it can only be proposed as initial clusters as in all past research (e.g., \citet{Yaakoubi2019,Yaakoubi2020}) and not as an initial solution. In this case, a generic standard initial solution is used, and since the initial solution and the initial clusters are different, an adaptation strategy is required to adapt the proposed clusters of the current window to the solution of the previous window. This is a major limitation of past research since the explored neighborhood needed to be large enough to reach a good LP (linear programming) solution but small enough to maintain a small number of fractional variables permitting to have an efficient heuristic branch-\&-bound. Thus, not only can we conclude that the primary metric of interest in our case is the feasibility of the constructed monthly solution. But, it also becomes crucial to propose a feasible pairing solution that can be proposed both as initial clusters and as an initial solution, therefore bypassing the adaptation strategy and permitting to reduce the resolution time (by reducing the neighborhood in which to explore in order to find a suitable solution).

Table~\ref{feasibility-table} summarizes the computational results on the feasibility and characteristics of constructed monthly pairings. First, breaking all illegal pairings in \textit{``Standard - initial solution'' } removes 50.56\% of the pairings, while that in \textit{``CNN - initial solution''} led to removing 21.05\%. Note that even though the test error is low for CNNs, due to the large number of infeasible pairings, running the optimization using this initial solution is problematic. Breaking all illegal pairings in \textit{``CNN-CRF - initial solution''} removes 12.12\% of the pairings, while only 11.07\% are removed from \textit{``Struct-CKN - initial solution''}. Clearly, although Struct-CKN has 97\% fewer parameters than other methods and its hyperparameters are not exhaustively fine-tuned, it outperforms other methods in terms of test error and feasibility. In what follows, we provide the constructed monthly pairings to the solver and compare the resulting solutions when using the baseline solution~\cite{desaulniers2020}, CNN~\cite{Yaakoubi2020} and the proposed approach (Struct-CKN).

\begin{table}
\centering
\caption{Characteristics of Monthly Solutions}
\resizebox{\linewidth}{!}{
\begin{tabular}{lccc}
\hline
                             &\#pairings& Cost    &     \% infeasible  \\
                             &          &  ($\times 10^8$)  &  pairings \\         
\hline
Standard - initial  & 6 525    & 37.93 & 50.56\\
Standard - feasible          &   3 226  & 29.75 &  \\
\hline
CNN - initial       &    4 883 & 24.13 & 21.05 \\
CNN - feasible               &     3 855& 16.58 & \\
\hline
CNN-CRF - initial   &   4 567  & 21.15 & 12.12 \\
CNN-CRF - feasible           &    4 010 & 16.15 & \\
\hline
Struct-CKN - initial& 4 515    & 20.29 & \textbf{11.07} \\
Struct-CKN - feasible        & \textbf{4 015}    & 16.46 & \\
\hline
\end{tabular}
}
\label{feasibility-table}
\end{table}

\subsubsection{Results on the Crew Pairing Problem}\label{sec:or_results}

As in Figure~\ref{fig:opt_process}, the baseline solution (\textit{``Baseline''}) for the monthly CPP is obtained by feeding the solver initial clusters from \textit{``Standard - feasible''}~\citep{desaulniers2020}. The previous state-of-the-art was obtained by feeding the solver initial clusters from \textit{``CNN - feasible''}, yielding a solution called \textit{``CNN''}. Instead of CNNs, we can use Struct-CKN to propose \textbf{both initial clusters and an initial solution} from \textit{``Struct-CKN - feasible''} to the solver, yielding a monthly solution called \textit{``Struct-CKN''}. To work on finding the best monthly solution possible and overcome the limitations of using the windowing approach, we feed the solution obtained from \textit{``Struct-CKN''} to the solver (again) as initial clusters and initial solution, yielding \textit{``Struct-CKN\textbf{+}''}. Because we cannot use the constructed monthly pairing as an initial solution, when using CNNs, we claim that re-running the optimization using the solution \textit{``CNN''} as initial clusters does not improve the monthly solution much. To support our claim, we feed solution \textit{``CNN''} to the solver as initial clusters, yielding \textit{``CNN\textbf{+}''} (see Figure~\ref{fig:opt_process}). See Appendix~\ref{appendix:cpp_opt_results} for detailed statistics of the optimization process.

Table~\ref{table-final-results} reports computational results for the final monthly solution.
Note that we do not report variances since the CPP solver is deterministic.
Struct-CKN outperforms both Baseline and CNN reducing the solution cost (in millions of dollars) and cost of global constraints by 9.51\% and 80.25\%, respectively, while also being 33\% faster than CNN. By re-running the optimization, on the one hand, CNN\textbf{+} does not improve the solution much. On the other hand, the Struct-CKN\textbf{+} solution yields the best statistics, reducing solution cost and cost of global constraints by 16.93\% and 97.24\%, respectively. More interestingly, the number of deadheads (see Section~\ref{sec:cpp}) is reduced by 41.23\%, compared to Baseline.
Therefore, we can conclude that proposing a feasible monthly solution both as initial clusters and as an initial solution allowed us to achieve better results in less time and to provide the possibility to re-optimize the solution and improve it further. Thus, this permits to update the training solutions, suggesting that Struct-CKN can be further optimized and that further research to avoid the windowing approach and use a one-month window can present better results than the current version of solver.

\begin{table}
\centering
\caption{Computational Results for Monthly Solutions}
\resizebox{\linewidth}{!}{
\begin{tabular}{lrrrr}
\hline
         & Solution  & Cost of global & Number of & Total\\
         &     cost      &  constraints & deadheads & time  \\
         & ($\times 10^6$ \textbf{\$}) &  ($\times 10^5$) & & (hours) \\
 \cline{1-1} \cline{2-5}
Baseline  & 20.64   & 21.27    & 992 &  45.92 \\
\citep{desaulniers2020} & & & & \\
\midrule
CNN~\citep{Yaakoubi2020} & 18.88      & 4.66      & 1014  & 95.72 \\
Struct-CKN       & \textbf{18.68}       & \textbf{4.20}      & \textbf{915} & 64.48 \\
\midrule
CNN\textbf{+} & 18.62   & 3.34  & 997 & 126.62 \\
Struct-CKN\textbf{+} & \textbf{17.15}       & \textbf{0.59}       & \textbf{583} & 41.44  \\
\hline
\end{tabular}
}
\label{table-final-results}
\end{table}

\section{Conclusion}
\label{sec:conclusion}
Seeking an initial solution to a crew pairing solver, this study proposes Struct-CKN, a new deep structured predictor. Its supervised use outperforms state-of-the-art methods in terms of primal sub-optimality of the structured prediction ``layer'' and test accuracy on the OCR dataset. The proposed method is then applied on a flight-connection dataset, modeled as a general CRF capable of incorporating local constraints in the learning process. To warm-start the solver, we use Struct-CKN to propose initial clusters and an initial solution to the solver, reducing the solution cost by 17\% (a gain of millions of dollars) and the cost of global constraints by 97\%, compared to baselines.
Future research will look into combining deep structured methods with various operations research methods and designing new reactive/learning metaheuristics that learn to guide the search for better solutions in real-time.

\section*{Acknowledgements}

We are thankful to the anonymous reviewers and the meta-reviewer for their valuable comments that improved the quality of this work. We also would like to thank Andjela Mladenovic, Gina Arena, Joey Bose, Mehdi Abbana Bennani, and Stephanie Cairns for their constructive comments regarding this work. We thank Iban Harlouchet for his involvement during the first phase of the project. This work was supported by IVADO, a Collaborative Research and Development Grant from the Natural Sciences and Engineering Research Council of Canada (NSERC), by the Canada CIFAR AI Chair Program, and AD OPT, a division of IBS Software. We would like to thank these organizations for their support and confidence. Simon Lacoste-Julien is a CIFAR Associate Fellow of the Learning in Machines \& Brains program.

\bibliography{bibliography}

\begin{thebibliography}{45}
\providecommand{\natexlab}[1]{#1}
\providecommand{\url}[1]{\texttt{#1}}
\expandafter\ifx\csname urlstyle\endcsname\relax
  \providecommand{\doi}[1]{doi: #1}\else
  \providecommand{\doi}{doi: \begingroup \urlstyle{rm}\Url}\fi

\bibitem[Belanger \& McCallum(2016)Belanger and McCallum]{belanger2016}
Belanger, D. and McCallum, A.
\newblock Structured prediction energy networks.
\newblock In \emph{International Conference on Machine Learning}, 2016.

\bibitem[Bengio et~al.(2020)Bengio, Lodi, and Prouvost]{bengio2020}
Bengio, Y., Lodi, A., and Prouvost, A.
\newblock Machine learning for combinatorial optimization: a methodological
  tour d’horizon.
\newblock \emph{European Journal of Operational Research}, 2020.

\bibitem[Bietti \& Mairal(2019)Bietti and Mairal]{Bietti2019}
Bietti, A. and Mairal, J.
\newblock Group invariance, stability to deformations, and complexity of deep
  convolutional representations.
\newblock \emph{JMLR}, 2019.

\bibitem[Bottou(2012)]{bottou2012}
Bottou, L.
\newblock Stochastic gradient descent tricks.
\newblock In \emph{Neural networks: Tricks of the trade}. Springer, 2012.

\bibitem[Buchta et~al.(2012)Buchta, Kober, Feinerer, and Hornik]{buchta2012}
Buchta, C., Kober, M., Feinerer, I., and Hornik, K.
\newblock Spherical k-means clustering.
\newblock \emph{Journal of Statistical Software}, 2012.

\bibitem[Chandra et~al.(2017)Chandra, Usunier, and Kokkinos]{chandra2017}
Chandra, S., Usunier, N., and Kokkinos, I.
\newblock Dense and low-rank gaussian crfs using deep embeddings.
\newblock In \emph{IEEE International Conference on Computer Vision}, 2017.

\bibitem[Chen et~al.(2015)Chen, Schwing, Yuille, and Urtasun]{chen2015}
Chen, L.-C., Schwing, A., Yuille, A., and Urtasun, R.
\newblock Learning deep structured models.
\newblock In \emph{International Conference on Machine Learning}, 2015.

\bibitem[Chu et~al.(2016)Chu, Ouyang, Li, and Wang]{CRFCNN}
Chu, X., Ouyang, W., Li, h., and Wang, X.
\newblock Crf-cnn: Modeling structured information in human pose estimation.
\newblock In \emph{Advances in Neural Information Processing Systems 29}, pp.\
  316--324. Curran Associates, Inc., 2016.

\bibitem[Desaulniers et~al.(1997)Desaulniers, Desrosiers, Dumas, Marc, Rioux,
  Solomon, and Soumis]{Desaulniers1997}
Desaulniers, G., Desrosiers, J., Dumas, Y., Marc, S., Rioux, B., Solomon,
  M.~M., and Soumis, F.
\newblock Crew pairing at {A}ir {F}rance.
\newblock \emph{European journal of operational research}, 97\penalty0 (2),
  1997.

\bibitem[Desaulniers et~al.(2020)Desaulniers, Lessard, Mohammed, and
  François]{desaulniers2020}
Desaulniers, G., Lessard, F., Mohammed, S., and François, S.
\newblock Dynamic constraint aggregation for solving very large-scale airline
  crew pairing problems.
\newblock \emph{Les Cahiers du GERAD}, G–2020\penalty0 (21), 2020.

\bibitem[Elhallaoui et~al.(2010)Elhallaoui, Metrane, Soumis, and
  Desaulniers]{Elhallaoui2008_multi}
Elhallaoui, I., Metrane, A., Soumis, F., and Desaulniers, G.
\newblock Multi-phase dynamic constraint aggregation for set partitioning type
  problems.
\newblock \emph{Mathematical Programming}, 123\penalty0 (2), 2010.

\bibitem[Gasse et~al.(2019)Gasse, Ch{\'e}telat, Ferroni, Charlin, and
  Lodi]{gasse2019}
Gasse, M., Ch{\'e}telat, D., Ferroni, N., Charlin, L., and Lodi, A.
\newblock Exact combinatorial optimization with graph convolutional neural
  networks.
\newblock In \emph{Advances in Neural Information Processing Systems}, 2019.

\bibitem[Graber et~al.(2018)Graber, Meshi, and Schwing]{graber2018}
Graber, C., Meshi, O., and Schwing, A.
\newblock Deep structured prediction with nonlinear output transformations.
\newblock In \emph{Advances in Neural Information Processing Systems}, 2018.

\bibitem[Greff et~al.(2016)Greff, Srivastava, Koutn{\'\i}k, Steunebrink, and
  Schmidhuber]{lstm}
Greff, K., Srivastava, R.~K., Koutn{\'\i}k, J., Steunebrink, B.~R., and
  Schmidhuber, J.
\newblock Lstm: A search space odyssey.
\newblock \emph{IEEE transactions on neural networks and learning systems},
  28\penalty0 (10), 2016.

\bibitem[Gygli et~al.(2017)Gygli, Norouzi, and Angelova]{gygli2017}
Gygli, M., Norouzi, M., and Angelova, A.
\newblock Deep value networks learn to evaluate and iteratively refine
  structured outputs.
\newblock \emph{International Conference on Machine Learning}, 2017.

\bibitem[Kasirzadeh et~al.(2017)Kasirzadeh, Saddoune, and
  Soumis]{kasirzadeh2017}
Kasirzadeh, A., Saddoune, M., and Soumis, F.
\newblock Airline crew scheduling: models, algorithms, and data sets.
\newblock \emph{EURO Journal on Transportation and Logistics}, 2017.

\bibitem[Khalil et~al.(2017)Khalil, Dai, Zhang, Dilkina, and Song]{khalil2017}
Khalil, E., Dai, H., Zhang, Y., Dilkina, B., and Song, L.
\newblock Learning combinatorial optimization algorithms over graphs.
\newblock In \emph{Advances in Neural Information Processing Systems}, 2017.

\bibitem[Kipf \& Welling(2016)Kipf and Welling]{Kipf}
Kipf, T.~N. and Welling, M.
\newblock Semi-supervised classification with graph convolutional networks.
\newblock \emph{International Conference on Learning Representations}, 2016.

\bibitem[Kumar et~al.(2005)Kumar, August, and Hebert]{Kumar2005}
Kumar, S., August, J., and Hebert, M.
\newblock Exploiting inference for approximate parameter learning in
  discriminative fields: An empirical study.
\newblock In \emph{International Workshop on Energy Minimization Methods in
  Computer Vision and Pattern Recognition}. Springer, 2005.

\bibitem[Lacoste-Julien et~al.(2012)Lacoste-Julien, Jaggi, Schmidt, and
  Pletscher]{lacoste2012}
Lacoste-Julien, S., Jaggi, M., Schmidt, M., and Pletscher, P.
\newblock Block-coordinate frank-wolfe optimization for structural svms.
\newblock \emph{International Conference on Machine Learning}, 2012.

\bibitem[Lacoste-Julien et~al.(2013)Lacoste-Julien, Jaggi, Schmidt, and
  Pletscher]{LJ-Jaggi2013}
Lacoste-Julien, S., Jaggi, M., Schmidt, M., and Pletscher, P.
\newblock Block-coordinate frank-wolfe optimization for structural svms.
\newblock \emph{International Conference on Machine Learning}, 2013.

\bibitem[Le~Priol et~al.(2018)Le~Priol, Pich{\'e}, and
  Lacoste-Julien]{Priol2018}
Le~Priol, R., Pich{\'e}, A., and Lacoste-Julien, S.
\newblock Adaptive stochastic dual coordinate ascent for conditional random
  fields.
\newblock \emph{Conference on Uncertainty in Artificial Intelligence}, 2018.

\bibitem[Leblond et~al.(2017)Leblond, Alayrac, Osokin, and
  Lacoste-Julien]{leblond2017}
Leblond, R., Alayrac, J.-B., Osokin, A., and Lacoste-Julien, S.
\newblock Searnn: Training rnns with global-local losses.
\newblock \emph{International Conference on Machine Learning}, 2017.

\bibitem[Mairal(2016)]{Mairal2016}
Mairal, J.
\newblock End-to-end kernel learning with supervised convolutional kernel
  networks.
\newblock In \emph{Advances in Neural Information Processing Systems}, 2016.

\bibitem[Mairal et~al.(2014)Mairal, Koniusz, Harchaoui, and Schmid]{Mairal2014}
Mairal, J., Koniusz, P., Harchaoui, Z., and Schmid, C.
\newblock Convolutional kernel networks.
\newblock \emph{Advances in Neural Information Processing Systems}, 2014.

\bibitem[Martins et~al.(2015)Martins, Figueiredo, Aguiar, Smith, and
  Xing]{martins2015}
Martins, A.~F., Figueiredo, M.~A., Aguiar, P.~M., Smith, N.~A., and Xing, E.~P.
\newblock Ad3: Alternating directions dual decomposition for map inference in
  graphical models.
\newblock \emph{Journal of Machine Learning Research}, 2015.

\bibitem[M{\"u}ller \& Behnke(2014)M{\"u}ller and Behnke]{pystruct}
M{\"u}ller, A.~C. and Behnke, S.
\newblock Pystruct: learning structured prediction in python.
\newblock \emph{J. Mach. Learn. Res.}, 15, 2014.

\bibitem[Murphy(2012)]{Murphy2012}
Murphy, K.~P.
\newblock \emph{Machine learning: a probabilistic perspective}.
\newblock MIT press, 2012.

\bibitem[Owerko et~al.(2020)Owerko, Gama, and Ribeiro]{owerko2020}
Owerko, D., Gama, F., and Ribeiro, A.
\newblock Optimal power flow using graph neural networks.
\newblock In \emph{ICASSP 2020-2020 IEEE International Conference on Acoustics,
  Speech and Signal Processing (ICASSP)}. IEEE, 2020.

\bibitem[Paszke et~al.(2019)Paszke, Gross, Massa, Lerer, Bradbury, Chanan,
  Killeen, Lin, Gimelshein, Antiga, Desmaison, Kopf, Yang, DeVito, Raison,
  Tejani, Chilamkurthy, Steiner, Fang, Bai, and Chintala]{pytorch}
Paszke, A., Gross, S., Massa, F., Lerer, A., Bradbury, J., Chanan, G., Killeen,
  T., Lin, Z., Gimelshein, N., Antiga, L., Desmaison, A., Kopf, A., Yang, E.,
  DeVito, Z., Raison, M., Tejani, A., Chilamkurthy, S., Steiner, B., Fang, L.,
  Bai, J., and Chintala, S.
\newblock Pytorch: An imperative style, high-performance deep learning library.
\newblock \emph{Advances in Neural Information Processing Systems 32}, 2019.

\bibitem[Pedregosa et~al.(2011)Pedregosa, Varoquaux, Gramfort, Michel, Thirion,
  Grisel, Blondel, Prettenhofer, Weiss, Dubourg, Vanderplas, Passos,
  Cournapeau, Brucher, Perrot, and Duchesnay]{scikit-learn}
Pedregosa, F., Varoquaux, G., Gramfort, A., Michel, V., Thirion, B., Grisel,
  O., Blondel, M., Prettenhofer, P., Weiss, R., Dubourg, V., Vanderplas, J.,
  Passos, A., Cournapeau, D., Brucher, M., Perrot, M., and Duchesnay, E.
\newblock Scikit-learn: Machine learning in {P}ython.
\newblock \emph{Journal of Machine Learning Research}, 12, 2011.

\bibitem[P{\'e}rez-Cruz et~al.(2007)P{\'e}rez-Cruz, Ghahramani, and
  Pontil]{perez2007}
P{\'e}rez-Cruz, F., Ghahramani, Z., and Pontil, M.
\newblock Kernel conditional graphical models.
\newblock In \emph{Predicting structured data}, 2007.

\bibitem[Scharstein \& Pal(2007)Scharstein and Pal]{Scharstein2007}
Scharstein, D. and Pal, C.
\newblock Learning conditional random fields for stereo.
\newblock In \emph{2007 IEEE Conference on Computer Vision and Pattern
  Recognition}. IEEE, 2007.

\bibitem[Schmidt et~al.(2015)Schmidt, Babanezhad, Ahmed, Defazio, Clifton, and
  Sarkar]{schmidt2015}
Schmidt, M., Babanezhad, R., Ahmed, M., Defazio, A., Clifton, A., and Sarkar,
  A.
\newblock Non-uniform stochastic average gradient method for training
  conditional random fields.
\newblock In \emph{AISTATS}, 2015.

\bibitem[Shalev-Shwartz \& Zhang(2013)Shalev-Shwartz and Zhang]{shalev2013b}
Shalev-Shwartz, S. and Zhang, T.
\newblock Stochastic dual coordinate ascent methods for regularized loss
  minimization.
\newblock \emph{Journal of Machine Learning Research}, 2013.

\bibitem[Shalev-Shwartz \& Zhang(2014)Shalev-Shwartz and Zhang]{shalev2014}
Shalev-Shwartz, S. and Zhang, T.
\newblock Accelerated proximal stochastic dual coordinate ascent for
  regularized loss minimization.
\newblock In \emph{International Conference on Machine Learning}, 2014.

\bibitem[Taskar et~al.(2004)Taskar, Guestrin, and Koller]{Taskar2004}
Taskar, B., Guestrin, C., and Koller, D.
\newblock Max-margin markov networks.
\newblock In \emph{Advances in Neural Information Processing Systems}, 2004.

\bibitem[Tran et~al.(2020)Tran, Garcez, Weyde, Yin, Zhang, and
  Karunanithi]{Tran2018}
Tran, S.~N., Garcez, A.~d., Weyde, T., Yin, J., Zhang, Q., and Karunanithi, M.
\newblock Sequence classification restricted boltzmann machines with gated
  units.
\newblock \emph{IEEE Transactions on Neural Networks and Learning Systems},
  2020.

\bibitem[Yaakoubi(2019)]{yaakoubi2019_thesis}
Yaakoubi, Y.
\newblock \emph{Combiner intelligence artificielle et programmation
  math{\'e}matique pour la planification des horaires des {\'e}quipages en
  transport a{\'e}rien}.
\newblock PhD thesis, Polytechnique Montr{\'e}al, 2019.

\bibitem[Yaakoubi et~al.(2019)Yaakoubi, Soumis, and
  Lacoste-Julien]{Yaakoubi2019}
Yaakoubi, Y., Soumis, F., and Lacoste-Julien, S.
\newblock Flight-connection prediction for airline crew scheduling to construct
  initial clusters for {OR} optimizer.
\newblock \emph{Les Cahiers du GERAD}, G–2019\penalty0 (26), 2019.

\bibitem[Yaakoubi et~al.(2020)Yaakoubi, Soumis, and
  Lacoste-Julien]{Yaakoubi2020}
Yaakoubi, Y., Soumis, F., and Lacoste-Julien, S.
\newblock Machine learning in airline crew pairing to construct initial
  clusters for dynamic constraint aggregation.
\newblock \emph{EURO Journal on Transportation and Logistics}, 2020.
\newblock ISSN 2192-4376.
\newblock \doi{10.1016/j.ejtl.2020.100020}.

\bibitem[Yen \& Birge(2006)Yen and Birge]{Birge2006}
Yen, J.~W. and Birge, J.~R.
\newblock A stochastic programming approach to the airline crew scheduling
  problem.
\newblock \emph{Transportation Science}, 40\penalty0 (1), 2006.

\bibitem[Zheng et~al.(2015{\natexlab{a}})Zheng, Jayasumana, Romera-Paredes,
  Vineet, Su, Du, Huang, and Torr]{CRFRNN}
Zheng, S., Jayasumana, S., Romera-Paredes, B., Vineet, V., Su, Z., Du, D.,
  Huang, C., and Torr, P.~H.
\newblock Conditional random fields as recurrent neural networks.
\newblock In \emph{Proceedings of the IEEE international conference on computer
  vision}, 2015{\natexlab{a}}.

\bibitem[Zheng et~al.(2015{\natexlab{b}})Zheng, Jayasumana, Romera-Paredes,
  Vineet, Su, Du, Huang, and Torr]{zheng2015}
Zheng, S., Jayasumana, S., Romera-Paredes, B., Vineet, V., Su, Z., Du, D.,
  Huang, C., and Torr, P.~H.
\newblock Conditional random fields as recurrent neural networks.
\newblock In \emph{Proceedings of the IEEE international conference on computer
  vision}, 2015{\natexlab{b}}.

\bibitem[Zhong \& Wang(2007)Zhong and Wang]{Zhong2007}
Zhong, P. and Wang, R.
\newblock Using combination of statistical models and multilevel structural
  information for detecting urban areas from a single gray-level image.
\newblock \emph{IEEE transactions on geoscience and remote sensing},
  45\penalty0 (5), 2007.

\end{thebibliography}
\bibliographystyle{icml2021}

\appendix

\onecolumn

The appendix is organized as follows:  we first provide details about the convolutional kernel networks (CKNs) and the multilayer convolutional kernel (MCK) in Appendix~\ref{appendix:ckn}. Next, we provide details on the Structured SVM approach in Appendix~\ref{appendix:struct_prediction}. Then, we present additional results on the OCR dataset in Appendix~\ref{appendix:add_sckn_results}. Then, we present the prediction problem formulation in Appendix~\ref{sec:app_prediction_problem_form} and describe the flight-connection dataset in Appendix~\ref{inst_descr_sec}. Finally, we provide details about the optimization process and instances used in our paper in Appendix~\ref{sec:opt_process} and present additional results for the Crew Pairing Problem (CPP) optimization in Appendix~\ref{appendix:cpp_opt_results}. Note that in the appendix, we will refer to Struct-CKN using CRF models as Struct-CKN-SDCA. Instead of CRFs, we can also use SSVM predictors, in particular the block-coordinate Frank-Wolfe (BCFW) algorithm, which we will refer to as Struct-CKN-BCFW.

\section{Convolutional Kernel Networks}\label{appendix:ckn}

This section contains additional details on CKNs~\citep{Mairal2014}.
We first present the Kernel trick, the reproducing kernel Hilbert space, and the Nystr\"om low-rank approximation in Sections~\ref{sec:kernel trick},~\ref{sec:rkhs}, and~\ref{sec:nystrom}, respectively. Then, Section~\ref{sec:mck} presents MCKs.

\subsection{Kernel Trick}\label{sec:kernel trick}

The kernel trick consists of embedding a ``raw input space'' $\mathcal{X}$ into a high dimensional Hilbert ``feature space'' $\mathcal{H}$, through a possibly non-linear mapping $\varphi: \mathcal{X} \to \mathcal{H}$, where the inner product in $\mathcal{H}$ admits the computation formula as in~\eqref{kerneltrick}, with $K: \mathcal{X} \times \mathcal{X} \mapsto \mathbb{R}$.

\begin{equation}
\label{kerneltrick}
\langle \varphi(x), \varphi(x')\rangle_\mathcal{H}=K(x,x'), \quad x, x' \in \mathcal{X}
\end{equation}

The ``kernel function'' $K$ is thought of as specifying similarity between elements of $\mathcal{X}$. In contrast, the similarities in $\mathcal{H}$ are expressed as simple inner products. 
The choice of an appropriate $\varphi$ can result in making the image of the data subset of $\mathcal{X}$ quasi-linearly separable, in the sense that the image of $\mathcal{X}$ and the image of another subset disjoint to $\mathcal{X}$ could (almost) be separated with a linear predictor.
When a linear predictor on $\mathcal{H}$ only uses inner products of $\mathcal{H}$ elements, then the formula (\ref{kerneltrick}) makes the training of this linear predictor computationally feasible.

The next section reviews a specific class of feature spaces that are built from positive definite kernels.

\subsection{Reproducing Kernel Hilbert Space}\label{sec:rkhs}

\begin{Definition} [\bfseries Positive definite kernels]
A positive definite (p.d.) kernel on a set $\mathcal{X}$ is a function $K : \mathcal{X}\times \mathcal{X} \to \mathbb{R}$ that is symmetric (i.e. $K(x,x')=K(x',x), \forall x,x'\in \mathcal{X}$), and for which, for any $N \in \mathbb{N}$ and any $(x_1, \ldots, x_N)\in \mathcal{X}^N$, the Gram matrix (or similarity matrix) $[K]_{ij}=K(x_i,x_j)$ is positive semi-definite (which means that for any $(a_1, \ldots, a_N) \in \mathbb{R}^n$, $\sum_{i=1}^{N}\sum_{j=1}^{N}a_ia_jK(x_i, x_j) \ge 0$).
\end{Definition}

\begin{Theorem}[\bfseries RKHS~\citep{Bietti2019}]
Let $K$ be a p.d. kernel on a set $\mathcal{X}$. There exists a unique Hilbert space $\mathcal{H} \subset \mathcal{X}^{\mathbb{R}}$ such that
\begin{enumerate}
\item $\{K: y \mapsto K(x,y), \mathcal{X} \mapsto \mathcal{R} \}_{x \in \mathcal{X}} \subset \mathcal{H}$
\item $\forall f \in \mathcal{H}, \forall x \in \mathcal{X}, f(x)=\langle f, K \rangle_\mathcal{H}$
\end{enumerate}
In particular, the application
\begin{align*}
\varphi :\mathcal{X} &\mapsto \mathcal{H} \\
x &\mapsto K
\end{align*}
which maps $\mathcal{X}$ in the ``feature space'' $\mathcal{H}$ satisfies
\begin{equation}
\langle \varphi(x), \varphi(x') \rangle_\mathcal{H} = K(x, x') \quad x,x' \in \mathcal{X}
\end{equation}
\end{Theorem}

We give now two classical examples of RHKSs associated to the euclidean space $\mathcal{X}=\mathbb{R}^d$: the linear kernel and the Gaussian kernel. The linear kernel on $\mathcal{X}$ is defined to be :
\begin{equation}
K_\text{lin}(x,x')=\langle x, x' \rangle_{\mathbb{R}^d}
\end{equation}
The RKHS associated to $(\mathcal{X}, K_{lin})$ is the Hilbert space
$\mathcal{H}=\left \{ f_w = \langle \bullet, w \rangle_{\mathbb{R}^d} \right\}_{w \in \mathbb{R}^d}$ endowed with the inner product
$ \langle f_w, f_v \rangle_{\mathcal{H}} = \langle w, v \rangle_{\mathbb{R}^d}$. This RKHS is isometric to $\mathbb{R}^d$. The Gaussian Kernel with bandwidth $\sigma$ on $\mathcal{X}$ is defined to be
\begin{equation}
K_\text{Gauss}(x,y)=e^{-\frac{\lVert x-y\rVert^2}{2\sigma^2}}
\end{equation}
The RKHS $\mathcal{H}$ associated to $(\mathcal{X}, K_{Gauss})$ is an infinite-dimensional feature space, and all points of $\mathcal{X}$ are mapped to the unit sphere of $\mathcal{H}$ ($\lVert \varphi(x) \rVert_\mathcal{H}^2=K(x,x)=1$).

\begin{Theorem}[\bfseries Representer theorem]
Let $\mathcal{X}$ be a set endowed with a p.d. kernel $K$, let $\mathcal{H}$ be its RKHS and $S=\{x_1, \ldots, x_n\} \subset \mathcal{X}$  a training data set. If $\Psi: \mathbb{R}^{n+1} \mapsto \mathbb{R}$ is strictly increasing with respect to the last variable, then 
\begin{align*}
f^* = & \argmin_{f\in \mathcal{H}}\Psi \left(f(x_1), \ldots, f(x_n), \lVert f \rVert_\mathcal{H} \right)\\
\implies & f^* \in \text{Span}\left (K_{x_1}, \ldots, K_{x_n} \right)
\end{align*}
\end{Theorem}

Remark: $f^*$ lives in a subspace of dimension $n$, although $\mathcal{H}$ can be of infinite dimension.

\subsection{Nystr\"om Low-rank Approximation}\label{sec:nystrom}

Consider a p.d. kernel $K:\mathcal{X} \times \mathcal{X} \mapsto \mathbb{R}$ and its RKHS $\mathcal{H}$, with the mapping $\varphi: \mathcal{X} \mapsto \mathcal{H}$ such that $K(x,x')=\langle \varphi(x), \varphi(x')\rangle_\mathcal{H}$. The Nystr\"om approximation consists in replacing any point $\varphi(x)$ in $\mathcal{H}$ by its orthogonal projection $\Pi_{\mathcal{F}}(x)$ onto a finite-dimensional subspace 
\begin{equation}
\mathcal{F} \defin \text{Span}(f_1, \ldots, f_p) \text{ with } p \ll n
\end{equation}
where the $f_i'$ are \textit{anchor points} in $\mathcal{H}$ (defined below).

This projetion is equivalent to 
\begin{equation}
\Pi_{\mathcal{F}}(x) \defin \sum_{j=1}^{p} \beta_j^*(x)f_j
\end{equation}
with
\begin{equation}
\beta^*(x) = \argmin_{\beta \in \mathbb{R}^p} \left \lVert \varphi(x)-\sum_{j=1}^{p}\beta_j f_j \right \rVert_\mathcal{H}^2
\end{equation}
Noting $[K_f]_{jl}=\langle f_j, f_l \rangle_\mathcal{H}$ and $f(x)=(f_1(x), \ldots, f_p(x))\in \mathbb{R}^p$, it is quickly checked that $\beta^*(x)=K_f^{-1}f(x)$
, where $K_f^{-1}$ is the inverse of the kernel matrix $K_f$ (or its pseudo-inverse, when the matrix $K_f$ is not full rank).
Thus,
\begin{equation}
\varphi(x) \approx \sum_{j=1}^p \beta_j^*(x)f_j
\end{equation}
and
\begin{equation}
\langle \varphi(x), \varphi(x') \rangle_\mathcal{H} \approx \beta^*(x)^\top K_f\beta^*(x')
\end{equation}
Defining the mapping 
\begin{align*}
\Gamma : \mathcal{X} & \mapsto \mathbb{R}^p \\
 x & \mapsto \Gamma(x)=K_f^{-1/2}f(x)
\end{align*}
as $\Gamma(x)=K_f^{1/2}\beta^*(x)$, we see that 
$$K(x,x') \approx \langle \Gamma(x), \Gamma(x') \rangle_{\mathbb{R}^d}$$
Consequently, the mapping $\varphi: \mathcal{X} \mapsto \mathcal{H}$ is approximated by a mapping $\Gamma : \mathcal{X}  \mapsto \mathbb{R}^p$ such that 
$$\langle \varphi(x), \varphi(x')\rangle_\mathcal{H} \approx \langle \Gamma(x), \Gamma(x') \rangle_{\mathbb{R}^d}$$
There exists various methods for choosing the anchor points $f_j$'s. When $\mathcal{X} = \mathbb{R}^d$, one such method consists in performing a $K$-means algorithm on a training data set $S=\{x_1, \ldots, x_n \} \subset \mathcal{X}$, to obtain $p$ centroids $\z_1, \ldots, \z_{p_1}\in \mathbb{R}^d$, and define anchor points as $f_j=\varphi(\z_j)$, for $j=1, \ldots, p$.

\subsection[Multilayer Convolutional Kernel]{Multilayer Convolutional Kernel~\citep{Mairal2016}}\label{sec:mck}

Two principal advantages of a CNN over a fully-connected neural network are: (i) sparsity - each nonlinear convolutional filter acts only on a local patch of the input, (ii) parameter sharing - the same filter is applied to each patch.

\begin{Definition}[\bfseries Image feature map and patch feature map]
An image feature map $\varphi$ is a function $\varphi : \Omega \to \mathcal{H}$, where $\Omega$ is a discrete subset of $[0,1]^2$ representing a set of pixel locations, and $\mathcal{H}$ is a Hilbert space representing a feature space. A patch feature map is the same as an image feature map with $\Omega$ replaced by a patch of pixels $\mathcal{P}$ centered at $0$.
\end{Definition}

\begin{Definition}[\bfseries Convolutional kernel]
 Let us consider two images represented by two image feature maps, respectively $\varphi$ and $\varphi':\Omega \to \mathcal{H}$. The one-layer convolutional kernel between $\varphi$ and $\varphi'$ is defined as 
\begin{equation}
\label{CKN1}
K(\varphi, \varphi')= \sum_{z \in \Omega}\sum_{z' \in \Omega} \left \{ \lVert \varphi(z) \rVert_{\mathcal{H}}\lVert \varphi'(z') \rVert_{\mathcal{H}}e^{-\frac{1}{2\beta^2}\lVert z-z'\rVert^2_2} \right. \nonumber \left. e^{-\frac{1}{2\sigma^2}\lVert \tilde{\varphi}(z)-\tilde{\varphi'}(z')\rVert^2_2} \right \}
\end{equation}
where $\tilde{\varphi}(z)=\left( 1/\lVert \varphi(z) \rVert_{\mathcal{H}}\right)\varphi(z)$ if $\varphi(z) \neq 0$ and $\tilde{\varphi}(z)=0$ otherwise. Idem for $\tilde{\varphi'}(z)$.
\end{Definition}

Remark 1: The role of $\beta$ is to control how much the kernel is locally shift-invariant.

Remark 2: $K$ is built on a kernel $k$ on $\mathcal{H}$, of the form 
\begin{equation}
k(h,h')=\lVert h \rVert_\mathcal{H}\lVert h' \rVert_\mathcal{H} \kappa\left( \frac{\langle h, h' \rangle}{\lVert h \rVert_\mathcal{H}\lVert h' \rVert_\mathcal{H}}\right), \quad h,h'\in \mathcal{H}
\end{equation}
According to a classical result (Schoenberg, 1942), when $\kappa$ is smooth with non-negative Taylor expansion coefficients, $k$ is a p.d. kernel. Consequently, $K$ is also a p.d. kernel.

\begin{Definition}[\bfseries Multilayer convolutional kernel]\label{def:multiscattering}
   Let us consider a discrete set of pixel locations $\Omega_{\kmone} \subseteq [0,1]^2$ and a Hilbert space $\mathcal{H}_{\kmone}$.
   We build a new set~$\Omega_k$ and a new Hilbert space~$\mathcal{H}_k$ as follows:

   (i) choose a patch shape~$\PP_k$ defined as a discrete symmetric subset
   of~$[-1,1]^2$, and a set of pixel locations~$\Omega_k$
   such that for all location~$\z_k$ in~$\Omega_k$, the patch~$\{\z_k\} + \PP_k$ is a subset of~$\Omega_{\kmone}$;
   In other words, each pixel location~$\z_k$ in~$\Omega_k$ corresponds to a valid patch of pixel locations in~$\Omega_{\kmone}$ centered at~$\z_k$.

   (ii) define the convolutional kernel~$K_k$ on the ``patch'' feature maps~$\PP_k \to
   \mathcal{H}_{\kmone}$, by replacing in~(\ref{CKN1}):~$\Omega$ by~$\PP_k$,~$\mathcal{H}$
   by~$\mathcal{H}_{\kmone}$, and~$\sigma,\beta$ by appropriate smoothing
   parameters~$\sigma_k,\beta_k$. We denote by~$\mathcal{H}_k$ the Hilbert space for
   which the positive definite kernel~$K_k$ is reproducing.

   An image represented by a feature map~$\varphi_{\kmone}: \Omega_{\kmone} \to
   \mathcal{H}_{\kmone}$ at layer~$\kmone$ is now encoded in the~$k$-th layer as~$\varphi_k:
   \Omega_k \to \mathcal{H}_k$, where for all~$\z_k$ in~$\Omega_k$,~$\varphi_{k}(\z_k)$ is the representation
   in~$\mathcal{H}_k$ of the patch feature map~$\z \mapsto \varphi_{\kmone}(\z_k + \z)$ for~$\z$ in~$\PP_k$.
\end{Definition}
In other words, given $\varphi_{\kmone}: \Omega_{\kmone} \to \mathcal{H}_{\kmone}$, we first define $\Omega_{k}$ and $\PP_k$ such that for any $\z_k \in \Omega_{k}$, $\z_k + \PP_k \subset \Omega_{\kmone}$. Second, we define $\mathcal{H}_{k}$ as being the RKHS of the set of patch feature maps
$$\{\z \mapsto \varphi(\z_k + \z), \PP_k \mapsto \mathcal{H}_{\kmone}: \varphi:\Omega_{\kmone} \to \mathcal{H}_{\kmone}, \z_k \in \Omega_k \}$$
endowed with the positive definite kernel $K_k$ adapted from (\ref{CKN1}), and we set $\varphi_{k}(\z_k)$ as being the representation
   in~$\mathcal{H}_k$ of the patch feature map~$\z \mapsto \varphi_{\kmone}(\z_k + \z)$. Thus we obtain $\varphi_{k}: \Omega_{k} \to \mathcal{H}_{k}$, and for any $\varphi_{\kmone}, \varphi'_{\kmone}: \Omega_{\kmone} \to \mathcal{H}_{\kmone}$, we have  $K_k \left(\varphi_{\kmone}, \varphi'_{\kmone} \right)= \langle \varphi_{k}, \varphi'_{k} \rangle_{\mathcal{H}_k}$.

\section{Graph-Based Learning}\label{appendix:graph_based_learning}

\subsection{Conditional Random Fields}\label{appendix:crf}

Algorithm~\ref{alg:sdca} describes this variant of SDCA, biasing the sampling towards examples with large duality gaps $g$. If we assume that the graph has a junction tree structure $\mathbb{T} = (\mathbb{C}, \mathbb{S})$, where $\mathbb{C}$ is the set of maximal cliques and $\mathbb{S}$ the set of separators, we can run message passing on the junction tree to infer new marginals given weights $w$: $ \hat{\mu}_i(w) = p (y_C = \cdot | x_i ; w) $, and recover joint probability $\alpha_i(y)$ as function of marginals $\mu_{i,C}$:
$\alpha_i(y) = \tfrac{ \prod_{C \in \mathbb{C}}^{} \mu_{i,C}(y_C) }{  \prod_{S \in \mathbb{S}}^{} \mu_{i,S}(y_S)}$.
This allows computing the entropy as in~\eqref{eq_entropy_sdca} and the Kullback-Leibler divergence of joints as in~\eqref{eq_duality_gaps_sdca}, using only the marginals $\mu_i$ and $\nu_i$ of $\alpha_i$ and $\beta_i$, respectively. With~\eqref{eq_entropy_sdca}, we compute the dual objective value, and perform the line search, as in~\eqref{eq_line_search_sdca}.

\begin{equation}
\widetilde{H}(\mu_i) = H(\alpha_i) = \sum_{C}{} H ( \mu_{i,C} ) - \sum_{S}{} H ( \mu_{i,S} )
\label{eq_entropy_sdca}
\end{equation}

\begin{equation}
\widetilde{D}(\mu_i || \nu_i) = \sum_{C}D_{KL}(\mu_{i,C} || \nu_{i,C}) -\sum_{S}D_{KL}(\mu_{i,S} || \nu_{i,S})
\label{eq_duality_gaps_sdca}
\end{equation}

\begin{equation}
\gamma^{*} = {\arg\!\max}_{\gamma \in [ 0,1  ]} \widetilde{H}( \mu_i^{(t)}  + \gamma \delta_i  ) - \frac{\lambda n}{2} \|  w^{(t)} + \gamma v_i \|^{2}
\label{eq_line_search_sdca}
\end{equation}

\begin{algorithm}[H]
\caption{SDCA for CRF~\citep{Priol2018}}\label{alg:sdca}
\begin{algorithmic}[1]
\STATE Initialize $\mu_i^{(0)} \in \Pi_C^{\Delta_C}, \forall i$; $w^{(0)} = \hat{w}(\mu^{(0)}) = \tfrac{1}{\lambda n}\sum_{i}B_i \mu_i^{(0)}$, where $B_i $ is the horizontal concatenation of the feature vectors over cliques
\STATE (Optional) Initialize the duality gaps $g_i = 100, \forall i$\
\FOR{t=1 : $n_{Ep\_SCDA}$}
\STATE Sample $i$ uniformly in \{1, \ldots , n\}
\STATE (Alternatively) Sample $i$ with $P(i) \propto g_i$
\STATE Let $\nu_{i,C}(y_C) = p( y_C | x_i ; w^{(t)} ) , \forall C \in \mathbb{C}$
\STATE (Optional) Let $g_i =  \tilde{D}(\mu_i || \nu_i)$~\eqref{eq_duality_gaps_sdca}
\STATE Let ascent direction be : {$ \delta_i = \nu_i - \mu_i^{t}$}, and the primal direction be : $v_i = \tfrac{1}{\lambda n} \hat{w}(\delta_i)$
\STATE Use line search to get the optimal step size~\eqref{eq_line_search_sdca}
\STATE Update $\mu_i^{(t+1)} = \mu_i^{(t)} + \gamma^{*} \delta_i$, and $w^{(t+1)} =  \hat{w}(\mu^{(t+1)}) = w^{(t)} + \gamma^{*} v_i$
\ENDFOR
\end{algorithmic}
\label{alg-sdca}
\end{algorithm}

Note that as in Section \ref{sec:sckn_framework}, to do inference for CRF models, we use AD3~\citep{martins2015} for approximate maximum a posteriori (MAP) inference. SDCA needs a marginalization oracle, whereas AD3 is used to do an approximate MAP.
In order to use SDCA with AD3, we propose a simple approximation, using MAP label estimates. We first run an approximate MAP inference algorithm and then imagine that the marginals put a unit probability at the approximate MAP solution and zero elsewhere~\citep{Kumar2005, Zhong2007, Scharstein2007}. This is a heuristic method, but it can be expected to work well when the estimated MAP solution is close to the true MAP, and the conditional distribution is strongly ``peaked''.

\subsection{Structural SVM - Block-Coordinate Frank-Wolfe}\label{appendix:struct_prediction}

To train CRFs, we either use a Structured SVM solver using BCFW algorithm~\citep{LJ-Jaggi2013}, or a stochastic optimization algorithm using SDCA~\citep{Priol2018}. In this Section, we briefly review SSVM.

Given a training set of input-output structure pairs $\{ (x_1,y_1), \ldots, (x_n,y_n) \}$ $\in$ $\mathcal{X} \times \mathcal{Y}$, a loss function $\Delta~:~\mathcal{Y} ~\times~\mathcal{Y}~\to \pmb{\mathbb{R}}$ and a joint feature map $F: \mathcal{X} \times \mathcal{Y} \to \pmb{\mathbb{R}} $ that encodes the input/output information necessary for prediction, we aim to learn $w$ so that the prediction function $f(x) = \argmax_{y} \langle w, F(x, y) \rangle$ minimizes the expected loss on future data.
To do so, we use SSVM predictors derived by maximum margin framework, and enforce correct output to be better than others by a margin: $ \langle w, F(x_n, y_n) \rangle \geq \Delta(y_n, y) + \langle w, F(x_n, y) \rangle ; ~ \forall ~ y \in \mathcal{Y}$.
This is a convex optimization problem, but non-differentiable, with many equivalent formulations, resulting in different training algorithms.
The Frank-Wolfe algorithm, one of such training algorithms, considers the convex minimization problem $\min_{\alpha \in M}f(\alpha )$, where $M$ is compact, and $f$ is continuously differentiable. It only requires optimizing linear functions over $M$. Since we experiment on large datasets, we use block-coordinate Frank-Wolfe (BCFW) algorithm~\citep{lacoste2012} showing better convergence than the original or the batch version.

\section {Additional results for the Struct-CKN predictor on the OCR dataset}\label{appendix:add_sckn_results}

\begin{figure}
   \centering
   \includegraphics[width=0.6\linewidth]{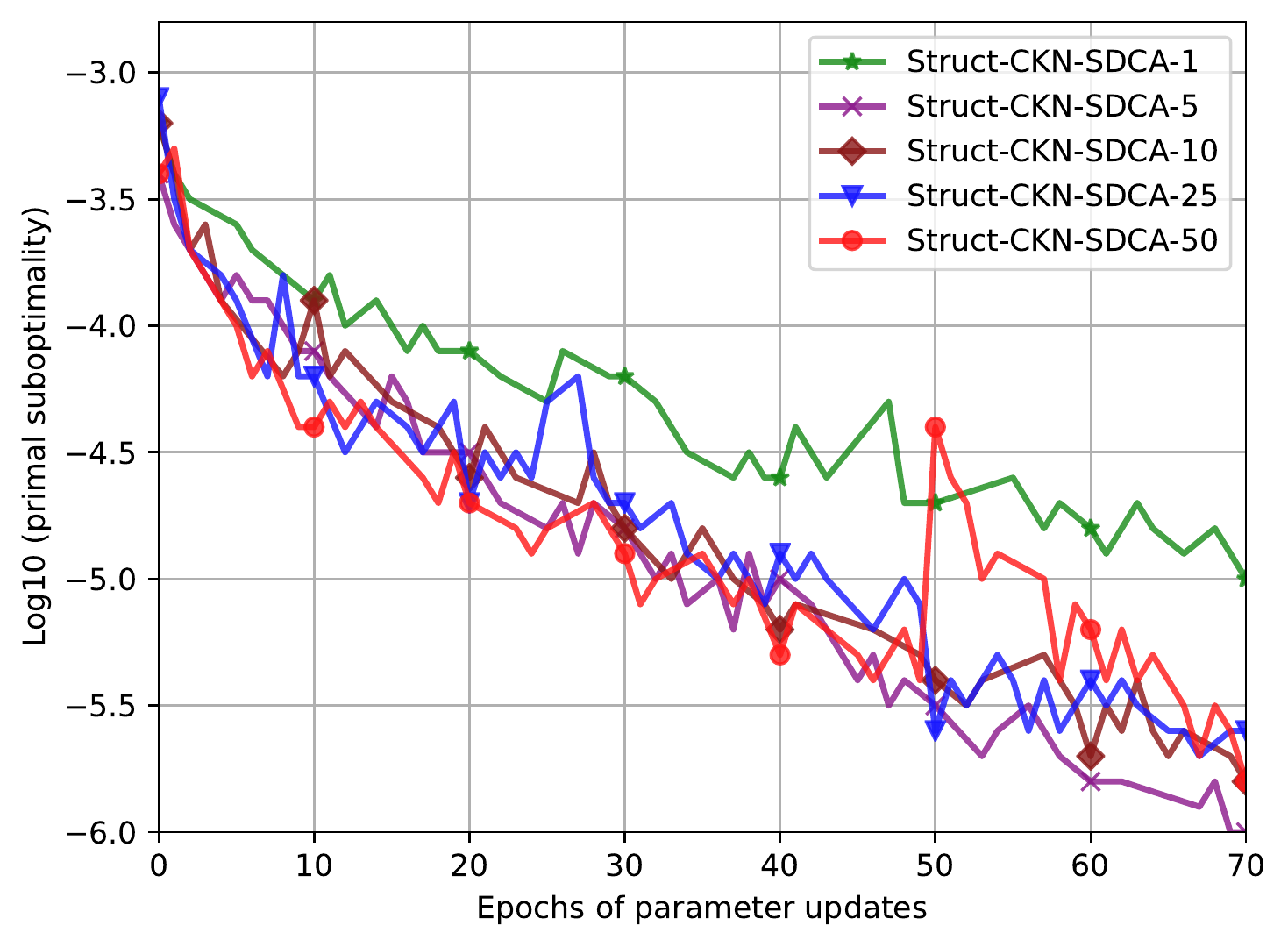}
   \caption{Struct-ckn With Gap Sampling With Various $n_{Ep\_SCDA}$ Values, as Indicated by the Number in the Legend}
   \label{fig:subopt_epochsSDCA}
\end{figure}

The hyperparameters for CKN are the number of layers, the number of filters and the size of patches. We use one layer and perform non-exhaustive grid search to set the number of filters (200) and the size of patches (5x5). We use cross-validation on the training set to measure the quality of the configuration and use different months for different samples to simulate the more realistic scenario in which we make a forecast over a new period.
To run SDCA, we use $\lambda = \tfrac{1}{n}$, where $n$ is the number of data points (sequences). As described in Algorithm~\ref{alg-sdca}, we use the gap sampling strategy: we sample proportionally to our current estimate of the duality gaps, with 80\% of uniform sampling. Finally, we use the Newton-Raphson algorithm on the derivative of the line search function.
We train the Struct-CKN-SDCA predictor using different values for $n_{Ep\_SCDA}$. As shown in Figure~\ref{fig:subopt_epochsSDCA}, we obtain similar results if we use values for $n_{Ep\_SCDA}$ between $5$ and $25$. Performing a single update of the SDCA parameter before updating the CKN weights makes the learning process difficult and using a value of 50 leads to the optimizer overfitting, making the learning process more difficult afterwards.

\begin{figure*}
\minipage{0.48\textwidth}
  \includegraphics[width=\linewidth]{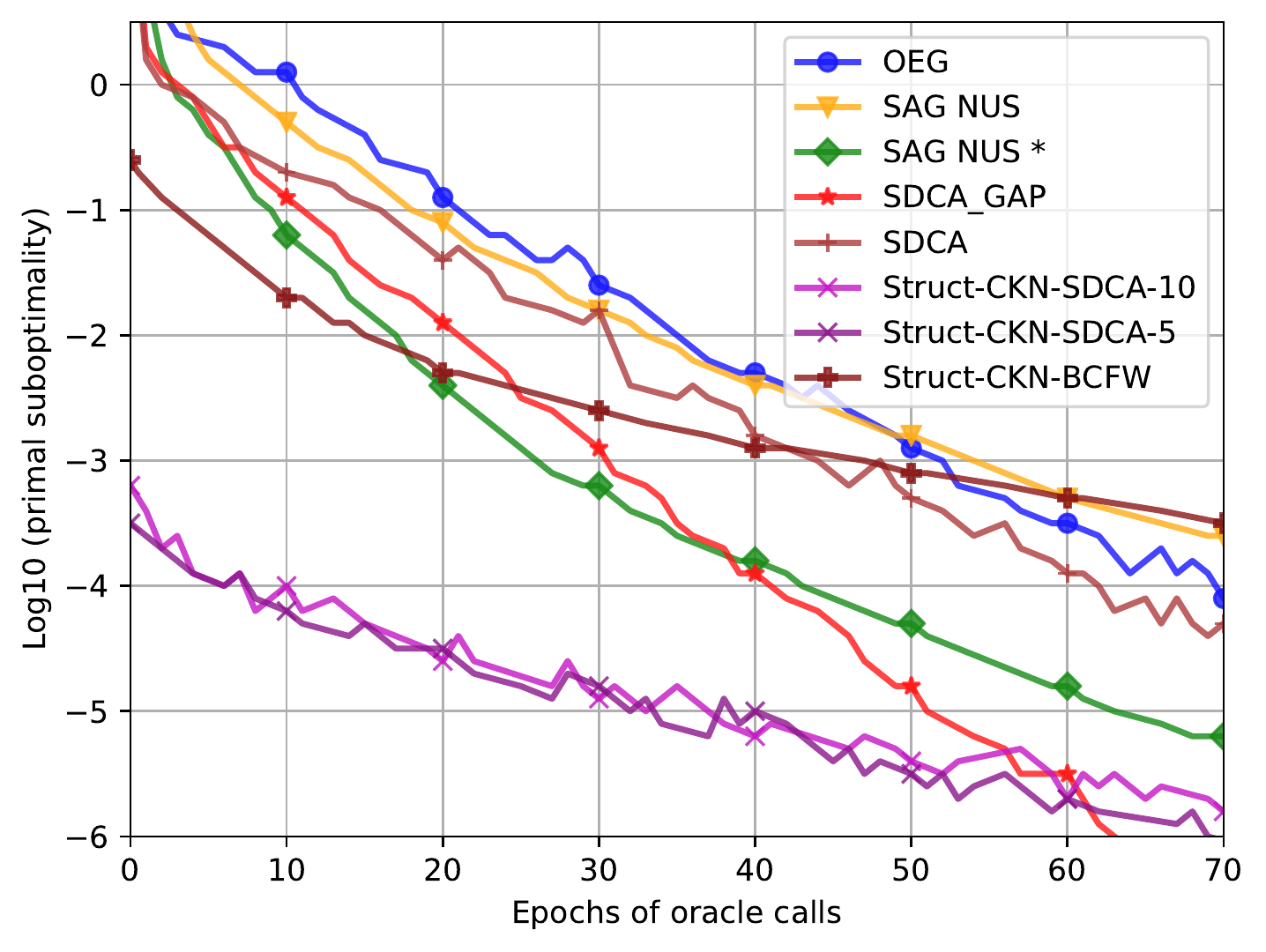}
  \label{fig:subopt_comparison_OrCall_appendix}
\endminipage\hfill
\minipage{0.48\textwidth}
  \includegraphics[width=\linewidth]{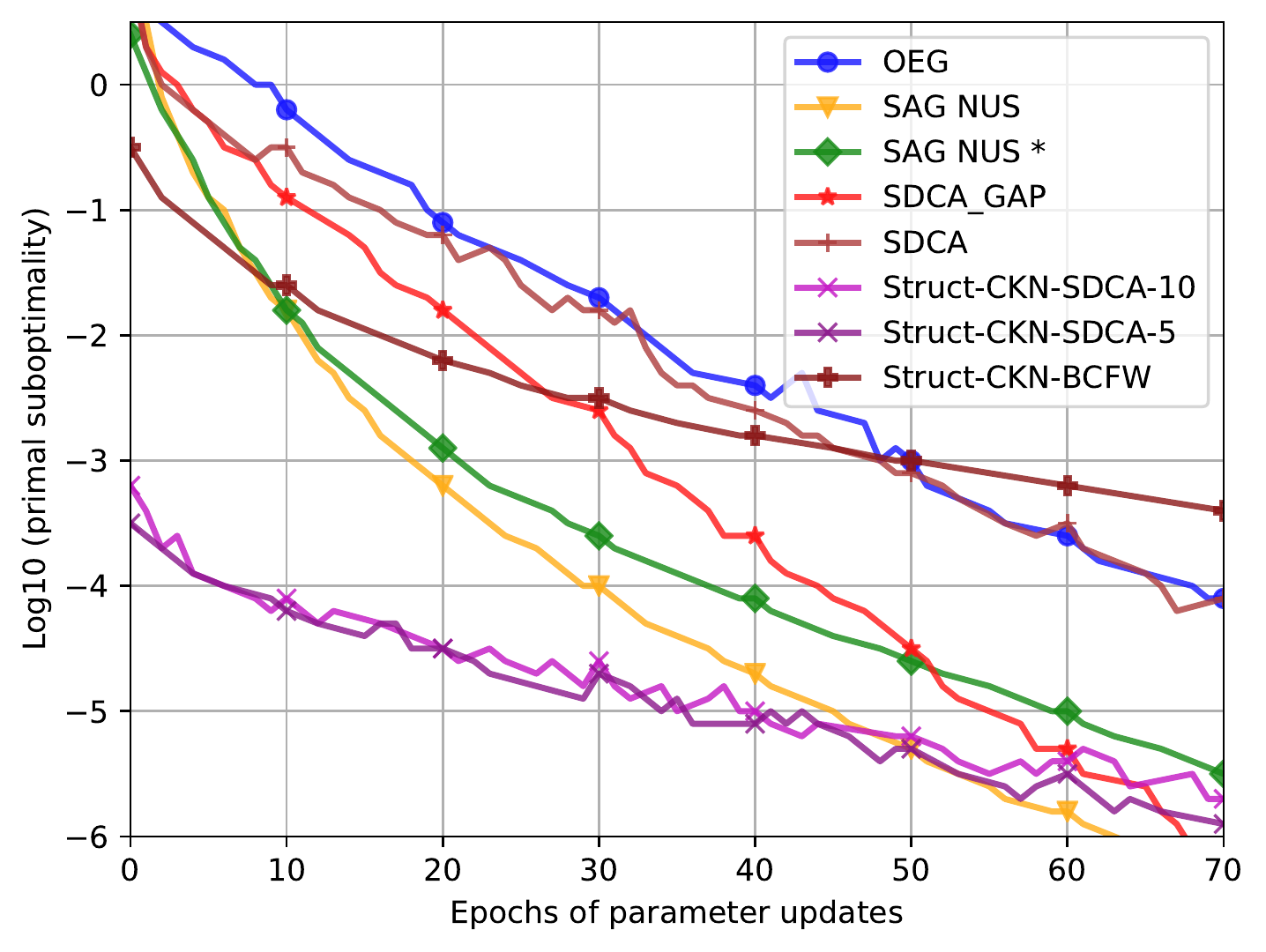}
  \label{fig:subopt_comparison_ParUpd_appendix}
\endminipage
\caption{Comparison of Primal Sub-optimality According to the Number of Oracle Calls (Left) or Parameter Updates (Right). SDCA Refers to Uniform Sampling. SDCA-GAP Refers to Sampling Gap Sampling 80\% of the Time. SAG-NUS Performs a Line Search at Every Iteration. SAG-NUS* Implements a Line-search Skipping Strategy. Struct-CKN-BCFW and Struct-CKN-SDCA Use BCFW and SDCA Respectively as Structured Predictors. Struct-CKN-SDCA-5 and Struct-CKN-SDCA-10 Use 5 and 10 Epochs Respectively to Train the SDCA Weights for Each Iteration.}
\label{fig:subopt_comparison_appendix}
\end{figure*}

\paragraph{Oracle Calls.}
\citet{schmidt2015} compared the algorithms based on the number of oracle calls. This metric was suitable for the methods they compared. Both OEG and SAG-NUS use a line search where they call an oracle on each step. SDCA, SDCA-GAP and Struct-CKN-SDCA do not require the oracle to perform their line search. However, the oracle is message passing on a junction tree. It has a cost proportional to the size of the marginals. Each iteration of the line search requires computing the entropy of these marginals, or their derivatives. These costs are roughly the same.

\paragraph{Parameter Updates.}
To give a different perspective, we also report the log of the sub-optimality against the number of parameter updates. This removes the additional cost of the line search for all methods.

\paragraph{Test Error on the OCR Dataset.}

We report in Table~\ref{table-test-ocr-appendix} the test errors on the OCR dataset.

\begin{table}
\centering
\caption{Test Error on the OCR Dataset}
\resizebox{\columnwidth}{!}{ 
\begin{tabular}{ccccccccccccc}
    \toprule
  &  SDCA - linear  &   LSTM   &  CNN-    & SCRBM  & Struct-CKN-    & Struct-CKN- \\
   &   features~\citep{Priol2018}    & ~\citep{lstm}       &  CRF~\citep{CRFCNN}    &~\citep{Tran2018}  &  BCFW    & SDCA  \\
    \midrule
    Test error (\%) & 12.0  &  4.6  &  4.5    & 4.03   &   3.42   &   \textbf{3.40}  \\
    \bottomrule
\end{tabular}
}
\label{table-test-ocr-appendix}
\end{table}

\section{Crew Pairing Problem}\label{appendix:cpp}

This section contains additional details on the Crew Pairing Problem (CPP).
We first describe the prediction problem formulation and the flight-connection dataset in Sections~\ref{sec:app_prediction_problem_form} and~\ref{inst_descr_sec}, respectively.
We describe the construction of the monthly crew pairing, once the ML predictor is trained, in Section \ref{sec:cnn_probs_monthly_cp}.
Then, we present in Section~\ref{sec:opt_process} in full the optimization process used to solve the CPP, once we obtain the initial solution and initial clusters using the ML predictor.
Finally, we report additional computational results for the CPP optimization per window and for monthly solution.

\subsection{Prediction Problem Formulation}\label{sec:app_prediction_problem_form}

The classification problem is the following: given the information about an incoming flight in a specific connecting city, choose among all the possible departing flights from this city (which can be restricted to the next 48 hours), the one that the crew should follow. These departing flights will be identified by their flight code (approximately 2,000 possible flight codes in the flight-connection dataset). Different flights may share the same flight codes in some airline companies, as flights performed multiple times a week usually use the same flight code. Nevertheless, a flight is uniquely identified by its flight code, departing city, and day, information that can be derived from the information on the incoming flight and the 48-hour window.

Each flight gives the following 5 features that we can use in our classification algorithm:
\begin{itemize}
\itemsep-0.1cm
    \item city of origin and city of destination ($\sim$200 choices);
    \item aircraft type (5 types);
    \item duration of flight (in minutes)
    \item arrival time (for an incoming flight)or departure time (for a departing flight).
\end{itemize}

Given that the aircrew arrived at a specific airport at a given time, we can use \emph{a priori} knowledge to define which flights are possible. For example, it is not possible to make a flight that starts ten minutes after the arrival, nor is it possible five days later. Furthermore, it is rare that the type of aircraft changes between flights since each aircrew is formed to use one or two types of aircraft at most. The reader is referred to~\citet{kasirzadeh2017} for further details on the likelihood of these scenarios. Note that even though we use the aircraft routing in the pre-processing steps, the classifier cannot know which next flight uses the same aircraft, as the prediction performance of the ML predictor is similar whether we use this information as a feature or not. The prediction problem is, therefore, not sensitive to this information and can do without it.

To construct the flight-connection dataset, the following conditions are used, which must always be met for the next flight performed by the crew:
\begin{itemize}
\itemsep-0.1mm
    \item The departure time of the next flight should follow the arrival time of the previous flight to the connecting city;
    \item The departure time of the next flight should not exceed 48 hours following the arrival time of the previous flight to the connecting city;
    \item The departure city of the next flight should be identical to the connecting city in the previous flight;
    \item The aircraft type should be the same. Indeed, crew scheduling is separable by crew category and aircraft type or family~\citep{kasirzadeh2017}.
\end{itemize}

For each incoming flight, all the departing flights in the next 48 hours are considered as a set. Then, only those that are feasible according to the masking constraints introduced in~\citet{Yaakoubi2019,yaakoubi2019_thesis} are considered to filter this set. The maximum number of possible flights is limited to 20, as it is sufficient in the airline industry, then the task is to predict the rank of the true label among this set of departing flights.

\subsection{Flight-connection Dataset}\label{inst_descr_sec}

Note that the instances in the flight-connection dataset~\citep{Yaakoubi2019} originate from an anonymous major airline and consist of six monthly crew pairing solutions for approximately 200 cities and approximately 50,000 flights per month. Each instance contains a one-month flight schedule, engine rotations, and crew pairings, as well as a list of airports and bases. This information is used to create input for the learning phase. The dataset consists of approximately 1,000 different source-destination pairs, 2,000 different flight codes, and pairings start from 7 different bases.

As shown in Figure~\ref{data-format}, this entry describes a pairing that takes place every Tuesday, Thursday, Friday, and Saturday between August 9, 2018 and September 3, 2018, except for the dates: 08/10, 08/20, and 08/29.
The pairing contains flights between Buffalo Niagara International Airport (BUF) and O'Hare International Airport (ORD). Apart from this, we know the distance between the two airports is 761 Km, from which we can extrapolate the duration of the flight, based on an average aircraft flight speed, which is equivalent to roughly 1 hour and 26 minutes.

\begin{figure}
\centerline{
\fbox{
\begin{minipage}{220pt}
O T 1821 0 CC 1 CAT 1 () "a" "" $\_2\_456\_$ 2018/08/09 2018/09/03 X 2018/08/10 2018/08/20 2018/08/29\\
\{\\
RPT ;\\
ABC05914~~ORD~1~02:10~BUF;\\
ABC06161~~BUF~1~09:50~ORD;\\
RLS;\\
\}\\
\end{minipage}
}
}
\caption{Data Format for Aircrew Pairings} \label{data-format}
\end{figure}

Note that the flight-connection dataset~\citep{Yaakoubi2019} contains an embedded representation of the flights' features and not the original features. An advantage of this approach is to anonymize further the dataset, an important condition to release the dataset in public domain. The dataset contains the output of the embedding layer, therefore an embedded representation of the input. For each incoming flight, the embedding layer (dimension $n_d$) is used to construct a feature representation for each of the 20 possible next flights. The concatenated matrix ($2D$ matrix) is a $n_d \times 20$ input, enabling the use of convolutional architecture across time. These representations were obtained during the last training epoch with a small tuned learning rate. This way, the representation of a city is not unique throughout the whole dataset. However, two representations of the same city are still similar.

Furthermore, note that to use Struct-CKN on the flight-connection dataset, we had to contact the authors of~\citet{Yaakoubi2019} in order to obtain the arcs (the connections between flights). This information is not part of the flight-connection dataset accessible in the public domain since disclosing it would easily permit the user to reconstruct the aeronautical map and identify the anonymous airline. The reader interested in reproducing our results on the flight-connection dataset is encouraged to contact authors of the paper or us. Finally, note that in contrast to the paper where the authors tackle a weekly crew pairing problem (approximately 10,000), and thus where the test set consists of the flights within a period of one week, our manuscript tackles a monthly crew pairing (50,000 flights/month). Therefore, although our test set is part of the flight-connection dataset, it is different from the test set considered in~\citet{Yaakoubi2019,yaakoubi2019_thesis}.

\subsection[From CNN Flight-connection Probabilities to Monthly Crew Pairings]{From CNN Flight-connection Probabilities to Monthly Crew Pairings~\citep{Yaakoubi2020}}\label{sec:cnn_probs_monthly_cp}

In what follows, we describe the construction of the monthly crew pairing once the CNN predictor is trained. As stated in Section~\ref{prediction_problem_form_sec}, we use the same methods and heuristics as in~\citet{Yaakoubi2020} in order to obtain \textit{``CNN - initial''}.

Upon the finalization of the flights-connection prediction CNN model training, we can use the same architecture to solve two other prediction problems on the test set (50,000 flights): (i) predict if each of the scheduled flights is the beginning of a pairing or not; and (ii) predict whether each flight is performed after a layover or not. In reality, the three predictors share the same representation. To solve these independent classification problems, we sum the three prediction problems' cross-entropy losses when learning, therefore performing a multi-output classification.

\paragraph{Start in a Base.} While training the CNN predictor to recognize whether a flight is the beginning of a pairing or not, it is possible that it misclassifies that a flight departing from a non-base city is indeed the beginning of a pairing. It is imperative to correct such false predictions in order to avoid pairings starting away from the base. Even though it is possible to construct a predictor to which only flights departing from the bases are given, it is more efficient and robust to use all flights in the training step. That way, the predictor learns a better representation of the input.

\paragraph{No Layover Below a Threshold.} While training the CNN predictor to recognize whether there is a layover or not between two flights, and since this decision is independent of finding the next flight, we can use a threshold on the number of hours between the previous and the next flight, below which it does not make sense to make a layover. This threshold should be defined considering previous solutions or by a practitioner.

To build the crew pairing, we use a greedy heuristic to build a crew pairing. Specifically, we consider each flight that the model predicts at the beginning of a pairing as a first flight. Given this incoming flight, we predict whether the crew is making a layover or not. In both cases, we consider the incoming flight and predict the next one. The pairing ends when the maximal number of days permitted per pairing is approached. We can use the above heuristic to construct a solution for the testing data, obtaining a monthly crew pairing that can be fed as initial clusters to the CPP solver. Unfortunately, if one flight in the pairing is poorly predicted, as the flights are predefined, the crew can finish its pairing away from the base. To correct the pairings, we define a heuristic in which all pairings where the crews finish away from the base are deleted.

Because the monthly problem is solved using a windowing approach with one-week windows and two days of overlap period, constructing an initial partition for the entire month and using the subset in each window to feed the solver can be a major flaw.
Such initial partition will have many inconsistencies with the solution of the previous window, particularly during the overlap period, such as flights belonging to two different clusters. We propose to adapt the proposed clusters to the solution of the previous window using the heuristics described above to construct the clusters of the current window in accordance with the solution found for the previous window and any inconsistency with the previous window is avoided, so the proposed partition is adapted to the current resolution. In addition, instead of only considering the flights that the model predicts at the beginning of a pairing as a first flight, incomplete clusters from the solution of the previous window starting during the overlap period are completed.

\subsection{Optimization Process}\label{sec:opt_process}

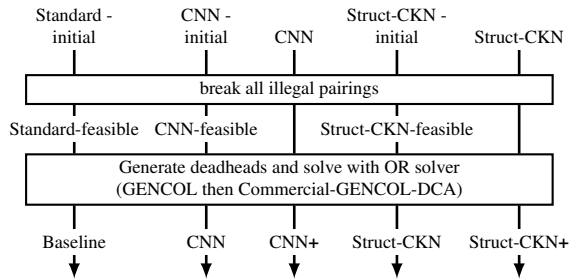
\begin{figure}
\centering
\begin{tikzpicture}[scale=0.65, every node/.style={scale=0.7}]

    \node[rectangle, minimum width=10cm, minimum height=0.5cm, draw, line width=0.8pt] (R1) at (0,0) {break all illegal pairings};
    \node[rectangle, minimum width=10cm, minimum height=0.5cm, align=center, draw, line width=0.8pt, anchor=north] (R2) at ([yshift=-1cm]R1.south) {Generate deadheads and solve with OR solver \\ (GENCOL then Commercial-GENCOL-DCA)};

    \draw[line width=1pt] ([xshift=1cm]R1.north west)--([yshift=0.5cm,xshift=1cm]R1.north west)node[above,align=center]{Standard -\\ initial};
    \draw[line width=1pt] ([xshift=1cm]R1.south west)--([xshift=1cm]R2.north west)node[midway,fill=white]{Standard-feasible};
    \draw[line width=1pt, -latex] ([xshift=1cm]R2.south west)--([xshift=1cm,yshift=-1.5cm]R2.south west)node[midway,fill=white]{Baseline};

    \draw[line width=1pt] ([xshift=-7.1cm]R1.north east)--([yshift=0.5cm,xshift=-7.1cm]R1.north east)node[above,align=center]{CNN -\\ initial};
    \draw[line width=1pt] ([xshift=-5.3cm]R1.north east)--([yshift=0.5cm,xshift=-5.3cm]R1.north east)node[above,align=center]{CNN};    
    \draw[line width=1pt] ([xshift=-3.2cm]R1.north east)--([yshift=0.5cm,xshift=-3.2cm]R1.north east)node[above,align=center]{Struct-CKN -\\ initial};
    \draw[line width=1pt] ([xshift=-7.1cm]R1.south east)--([xshift=-7.1cm]R2.north east)node[midway,fill=white]{CNN-feasible};
    \draw[line width=1pt, -latex] ([xshift=-7.1cm]R2.south east)--([xshift=-7.1cm,yshift=-1.5cm]R2.south east)node[midway,fill=white]{CNN};
    \draw[line width=1pt] ([xshift=-3.2cm]R1.south east)--([xshift=-3.2cm]R2.north east)node[midway,fill=white]{Struct-CKN-feasible};
    \draw[line width=1pt, -latex] ([xshift=-3.2cm]R2.south east)--([xshift=-3.2cm,yshift=-1.5cm]R2.south east)node[midway,fill=white]{Struct-CKN};
    
    \draw[line width=1pt] ([xshift=-0.7cm]R1.north east)--([yshift=0.5cm,xshift=-0.7cm]R1.north east)node[above]{Struct-CKN};
    \draw[line width=1pt] ([xshift=-0.7cm]R1.south east)--([xshift=-0.7cm]R2.north east);
    \draw[line width=1pt] ([xshift=-5.3cm]R1.south east)--([xshift=-5.3cm]R2.north east);
    \draw[line width=1pt, -latex] ([xshift=-0.7cm]R2.south east)--([xshift=-0.7cm,yshift=-1.5cm]R2.south east)node[midway,fill=white]{Struct-CKN\textbf{+}};
    \draw[line width=1pt, -latex] ([xshift=-5.3cm]R2.south east)--([xshift=-5.3cm,yshift=-1.5cm]R2.south east)node[midway,fill=white]{CNN\textbf{+}};

\end{tikzpicture}
\caption{The Optimization Process for the CPP}
\label{fig:opt_process_appendix}
\end{figure}

As shown in Figure \ref{fig:opt_process_appendix}, upon the finalization of training the Struct-CKN predictor weights, we build a monthly solution that we can provide to the optimization process as an initial solution and as initial clusters. Indeed, the optimization process requires a standard initial solution that is usually a ``cyclic'' weekly solution rolled to cover the whole month. In this monthly solution, pairings containing flights that have disappeared or have changed hours are broken; new pairings are then built to cover these flights as well, along with new flights that have appeared.

Instead of using a standard initial solution called \textit{``Standard - initial''}, we use the Struct-CKN predictor to construct the monthly solution.
One can use a greedy heuristic to build the crew pairing. Specifically, we consider each flight that the model predicts at the beginning of a pairing as a first flight. Given this incoming flight, we predict the next one. The pairing ends when the maximal number of days permitted per pairing is approached. We can use the above heuristic to construct a solution for test data, obtaining a monthly solution that can be fed both as an initial solution and as an initial partition to the solver.
Unfortunately, if one flight in the pairing is poorly predicted, the crew can finish its pairing away from the base. Therefore, for all pairings where crews finish away from the base, we delete all flights performed after the last time that the crew arrives at the base. Note that pairings may begin before the bidding period to cover all active flights. Therefore, using the Struct-CKN predictor, we propose pairings on an extended period to cover the maximal number of flights. This leads to a monthly crew pairing schedule that we consider as an initial solution called \textit{``Struct-CKN - initial - extended period''}. We can also restrict the pairings to be in the bid period. This leads to another monthly crew pairing that can also considered as an initial solution called \textit{``Struct-CKN - initial - bid period''}. Note that we only consider \textit{``Struct-CKN - initial''} in the main paper.
Then, we freeze the legal sub-part of pairings contained in the initial crew pairings by breaking all illegal pairings. Note that pairings starting before the bidding period are considered illegal since we would like to prevent the solver from generating pairings before the bid period starts. By breaking all illegal pairings from \textit{``Standard - initial''}, we obtain a monthly solution called \textit{``Standard - feasible''}. Similarly, by breaking all illegal pairings either from \textit{``Struct-CKN - initial - extended period''} or \textit{``Struct-CKN - initial - bid period''}, we obtain the same monthly solution called \textit{``Struct-CKN - feasible''}.

To cover the flights that just became uncovered, we generate deadheads on all flights and use the GENCOL solver to optimize with legal sub-blocks imposed covering open (active) flights. In this approach, the problem is solved by a ``rolling time horizon'' approach. Because the GENCOL solver is able to solve up to a few thousand flights per window, we are constrained to use two-days windows. This means that the month is divided into overlapping time slices of equal length. Then, a solution is constructed greedily in chronological order by solving the problem \emph{restricted} to each time slice sequentially, taking into account the solution of the previous slice through additional constraints.
The monthly solution obtained is then passed through to the CPP solver. The clusters can either be extracted from this solution (as is usually done) or given separately (as is the case in our approach). Therefore, by providing an initial solution, we can not only expedite the optimization process, calculate the number of pairings considered legal, the number of flights covered but also propose clusters that are similar to the initial solution, thus reducing the degree of incompatibility between the current solution and the proposed pairings.
Using the standard approach, we propose \textit{``Standard - feasible''} as an initial solution to GENCOL and obtain a solution called \textit{``GENCOL init.''}. Similarly, using the Struct-CKN predictor, we propose \textit{``Struct-CKN - feasible''} as an initial solution to GENCOL and obtain a solution called \textit{``Struct-CKN \textbf{+} GENCOL''}.

\subsection{Additional Results for The Crew Pairing Problem Optimization}\label{appendix:cpp_opt_results}

In what follows, we include additional results on the flight-connections dataset, and the characteristics of the monthly solutions constructed using probabilities yielded by different predictors in Table~\ref{table:test_acc_fcd_appendix} and Table~\ref{feasibility_table_appendix}, respectively. By using the monthly crew pairings to start the CPP solver, we report in Table~\ref{table_solver_results_appendix} and~\ref{table_final_results_appendix} the computational results per window for the optimization process, and computation results for the final monthly solution, respectively.

We report in Table \ref{table:test_acc_fcd_appendix} the test error on the flight-connection dataset using (1) CNNs and Gaussian Process (GP) to search for the best configuration of hyperparameters, as in~\citet{Yaakoubi2020}, (2) a standard CNN-CRF (with non-exhaustive hyperparameter tuning), and (3) our proposed predictor Struct-CKN. We obtain a test error of 0.32\%, 0.38\%, 0.35\% and 0.28\% using CNN, CNN-CRF, Struct-CKN-BCFW and Struct-CKN-SDCA, respectively. Therefore, our predictor outperforms both CNN (using hundreds of iterations of GP) and CNN-CRF. Note that although one can use GP to fine-tune hyperparameters, this is not feasible in a real-case usage scenario as practitioners cannot perform GP, each time new data is available.

Computational results on the feasibility and characteristics of initial solutions are summarized in Table \ref{feasibility_table_appendix}. We report the number of covered flights, pairings, and the cost of the monthly solution (i.e. cost of undercovered flights + cost of overcovered flights + solution cost of the covered flights). Note that the cost of the monthly solution is inflated and does not represent the real cost of the solution because the cost of undercovered and overcovered flights is expressed in large values. First, note that once we break all illegal pairings in \textit{``Standard - initial'' }, 51\% of the pairings and 45\% of the covered flights are removed, while we only remove 11\% of the pairings and 6\% of the covered flights for \textit{``Struct-CKN - initial'' }.

Table \ref{table_solver_results_appendix} reports the mean of computational results per window for all algorithms, namely, GENCOL-DCA, CNN, Struct-CKN and Struct-CKN\textbf{+}. For each window and each algorithm, we provide the LP value at the root node of the search tree N0 (LP-N0), computational time at N0 (N0 time), the number of fractional variables (\# VF-N0) in the current MP solution at N0, number of branching nodes resolved (\# Nodes), best LP value found (Best-LP), pairing cost of the best feasible solution (INT) and total computational time (T time); all times are in seconds.

While CNN gives better LP-N0, Best-LP, and INT values with an average reduction factor of 14.37\%, 14.17\% and 11.48\% respectively, compared to GENCOL-DCA, Struct-CKN yields better results with a reduction factor of 32.05\%, 32\% and 30.94\% respectively.
On the other hand, note that whilst CNN has a computational time on average 108\% larger than that of GENCOL-DCA, Struct-CKN has a computational time only 40\% larger on average.
This time increase is due to the number of fractional variables at N0 with an increase factor of 81\% for CNN and 18\% for Struct-CKN, compared to GENCOL-DCA.
When base constraints are restrictive, the root node solutions contain a more significant number of fractional-valued pairing variables in order to split the worked time between the bases evenly. This causes an increase in the number of branching nodes required to obtain a good integer solution.
While the time increase in CNN is due to the large number of fractional variables at N0, caused by inconsistencies between the initial solution and initial clusters, Struct-CKN has a lower number of nodes and computational time than CNN for two reasons. First, the prediction process is not a greedy process, in the sense that the links (connections) between flights are predicted jointly. This makes the solution more feasible, as shown in Section \ref{sec:feasibility}. Second, and unlike NN-based heuristics, we use the proposed pairings not only as initial clusters, but also as an initial solution. Next, we compare GENCOL-DCA and Struct-CKN\textbf{+}. Struct-CKN\textbf{+} gives better LP-N0 and Best-LP values for all windows providing an average reduction factor of 56.31\% and 55.73\% respectively. Likewise, Struct-CKN\textbf{+} gives better feasible solutions with a reduction factor of 56.26\%. This is explained by the ability of Struct-CKN\textbf{+} to propose a better initial solution, better than those used as training set for the Struct-CKN predictor.

Computational results for the monthly solution obtained at the end of the optimization process are reported in Table \ref{table-final-results}. We report the total solution cost, cost of global constraints, and number of deadheads obtained with Commercial-GENCOL-DCA fed by GENCOL init (baseline), CNN, CNN\textbf{+}, Struct-CKN, and Struct-CKN\textbf{+}.

While CNN reduces the solution cost and the cost of global constraints by 8.52\% and 78.11\%, Struct-CKN outperforms CNN reducing the solution cost and the cost of global constraints by 9.51\% and 80.25\%. Furthermore, while Struct-CKN reduces the number of deadheads in the solution by 7.76\%, note that for CNN, the number of deadheads in the solution is slightly larger with an increase factor of 2.17\%, compared to GENCOL-DCA. The cost of deadheads is accounted for in the solution cost. Because the solution cost is a multi-objective function, We believe that using slightly more deadheads permitted to get better solutions, enhancing both the solution cost and the cost of global constraints. The solutions found by Struct-CKN\textbf{+} present better statistics than GENCOL-DCA, CNN and Struct-CKN, with a reduction factor in solution cost and cost of global constraints of 16.93\% and 97.24\%, respectively. Even more interesting, the number of deadheads used is reduced by 41.23\%, compared to GENCOL-DCA, which shows that the Struct-CKN monthly solution can further be optimized and that further research to avoid the windowing approach and use a one-month window can present better results than the current version of Commercial-GENCOL-DCA.

\begin{table}[H]
\centering
\caption{Test Error on the Flight-connection Dataset}
\begin{tabular}{lcr}
\toprule
& Test error (\%) & \# parameters \\
\midrule
CNN~\citep{Yaakoubi2020} & 0.32 & 459 542 \\
\hline
CNN-CRF~\citep{CRFCNN} & 0.38 & 547 542 \\
\hline
Struct-CKN-BCFW &  0.35 & \textbf{15 200} \\
\hline
Struct-CKN-SDCA & \textbf{0.28} & \textbf{15 200} \\
\bottomrule
\end{tabular}
\label{table:test_acc_fcd_appendix}
\end{table}

\begin{table}[H]
\centering
\caption{Characteristics of Monthly Solutions}
\resizebox{\textwidth}{!}{
\begin{tabular}{cC{0.1mm}rrrrrrr}
\hline
                                                && Flight & \#undercovered      & \#overcovered      &  \#covered  & \#pairings  & Cost                     & \#deadheads \\
                                                && time   &    flights          &    flights         &   flights &   & (under. + over. + sol. ) &             \\
\cline{1-1} \cline{3-9}
Standard - initial                     && 81804h36    & 4 960         & 3 114        & 45 929  & 6 525        & 37 927 328 596                   & 475         \\
Standard - feasible          && 45644h21    & 25 838        & 24          & 25 051  & 3 226        & 29 753 057 875                   & 92          \\
GENCOL init.                                    && 82905h03    & 4 304         & 23          & 46 585  & 6 951        & 9 199 014 453                    & 1 718        \\
\cline{1-1} \cline{3-9}
Struct-CKN - initial - extended period && 81503h33    & 5 100         & 614         & 45 789  & 4 916        & 15 893 302 291                   & 266         \\
Struct-CKN - initial - bid period   && 75292h22    & 8 535         & 599         & 42 354  & 4 515        & 20 294 606 643                   &  241        \\
Struct-CKN - feasible        && 70672h27    & 11 148        & \textbf{0}           & 39 741  & \textbf{4 015}        & 16 45 833 4260                   & 220         \\
Struct-CKN \textbf{+} GENCOL                             && 82886h49    & 4 313         & 23          & 46 576  & \textbf{5 993}        & \textbf{9 198 925 708}                    & \textbf{1360}        \\
\hline
\end{tabular}}
\label{feasibility_table_appendix}
\end{table}

\begin{table}[H]
\centering
\caption{Computational Results per Window}
\resizebox{\textwidth}{!}{
\begin{tabular}{ccC{0.1mm}rrC{0.1mm}rC{0.1mm}R{1.3cm}C{0.1mm}rrC{0.1mm}rrC{0.1mm}r}
\hline
Win.&Alg. & ~ & LP-N0&Diff. & ~ &\#FV-N0& ~ &\#Nodes & ~ & Best-LP&Diff.    & ~ & INT&Diff. & ~ & T time\\
 & & ~ & & (\%) & ~ & & ~ & &  ~ & & (\%)& ~ &  &(\%)& ~ &(s) \\
\cline{1-2}  \cline{4-5}  \cline{7-7}  \cline{9-9} \cline{11-12} \cline{14-15}  \cline{17-17} 
       & GENCOL-DCA &  & 10 122 035 &        &  & 2719 &  & 159   &  & 10 025 487.73 &        &  & 10 276 344.14 &        &  & \textbf{7 891}      \\
       & CNN &  & 9 904 828  & -2.15  &  & 2870 &  & 203   &  & 9 891 390.17  & -1.34  &  & 10 136 106.73 & -1.36  &  & 10 691     \\
1      & Struct-CKN       &  & 7 099 264  & -29.86 &  & 2929 &  & 200   &  & 7 094 683.52  & -29.23 &  & 7 298 476.40  & -28.98 &  & 14 855     \\
       & Struct-CKN\textbf{+}       &  & \textbf{2 432 353}  & \textbf{-75.97} &  & \textbf{2413} &  & \textbf{170}   &  & \textbf{2 423 143.75}  & \textbf{-75.83} &  & \textbf{2 501 388.70}  & \textbf{-75.66} &  & 14 503     \\
       &          &  &          &        &  &         &  &      &  &             &        &  &             &        &  &            \\
       & GENCOL-DCA &  & 11 396 498 &        &  & 2519 &  & 162   &  & 11 225 022.17 &        &  & 11 501 016.42 &        &  & \textbf{27 778}     \\
       & CNN &  & 10 319 719 & -9.45  &  & 5848 &  & 419   &  & 10 215 252.88 & -9.00  &  & 10 865 255.80 & -5.53  &  & 94 004     \\
2      & Struct-CKN       &  & 7 547 443  & -33.77 &  & 3134 &  & 340   &  & 7 430 486.16  & -33.80 &  & 7 915 983.69  & -31.17 &  & 41 890     \\
       & Struct-CKN\textbf{+}       &  & \textbf{4 533 093}  & \textbf{-60.22} &  & \textbf{2011} &  & \textbf{113}   &  & \textbf{4 518 960.16}  & \textbf{-59.74} &  & \textbf{4 595 477.05}  & \textbf{-60.04} &  & 28 210     \\
       &          &  &          &        &  &         &  &      &  &             &        &  &             &        &  &            \\
       & GENCOL-DCA &  & 10 740 127 &        &  & 2330 &  & 177   &  & 10 641 496.00 &        &  & 11 107 990.12 &        &  & 30 421     \\
       & CNN &  & 9 596 326  & -10.65 &  & 4838 &  & 294   &  & 9 432 461.69  & -11.36 &  & 10 051 850.40 & -9.51  &  & 53 736     \\
3      & Struct-CKN       &  & 7 129 557  & -33.62 &  & 3045 &  & 198   &  & 6 977 006.00  & -34.44 &  & 7 400 639.68  & -33.38 &  & 34 255     \\
       & Struct-CKN\textbf{+}       &  & \textbf{4 588 242}  & \textbf{-57.28} &  & \textbf{1876} &  & \textbf{83}    &  & \textbf{4 571 315.93}  & \textbf{-57.04} &  & \textbf{4 650 601.49}  & \textbf{-58.13} &  & \textbf{27 003}     \\
       &          &  &          &        &  &         &  &      &  &             &        &  &             &        &  &            \\
       & GENCOL-DCA &  & 9 764 727  &        &  & 2606 &  & 229   &  & 9 591 592.13  &        &  & 9 968 081.73  &        &  & 33 898     \\
       & CNN &  & 8 063 084  & -17.43 &  & 4763 &  & 394   &  & 7 868 177.74  & -17.97 &  & 8 791 799.94  & -11.80 &  & 56 291     \\
4      & Struct-CKN       &  & 6 427 422  & -34.18 &  & 3387 &  & 183   &  & 6 291 641.09  & -34.40 &  & 6 439 030.07  & -35.40 &  & 45 430     \\
       & Struct-CKN\textbf{+}       &  & \textbf{4 617 230}  & \textbf{-52.72} &  & \textbf{1938} &  & \textbf{134}   &  & \textbf{4 601 886.81}  & \textbf{-52.02} &  & \textbf{4 697 567.30}  & \textbf{-52.87} &  & \textbf{27 737}     \\
       &          &  &          &        &  &         &  &      &  &             &        &  &             &        &  &            \\
       & GENCOL-DCA &  & 8 095 063  &        &  & 3150 &  & 188   &  & 7 899 149.01  &        &  & 8 102 703.93  &        &  & 35 948     \\
       & CNN &  & 6 076 461  & -24.94 &  & 5333 &  & 417   &  & 5 953 932.68  & -24.63 &  & 6 453 605.80  & -20.35 &  & 74 622     \\
5      & Struct-CKN       &  & 5 383 225  & -33.50 &  & 3038 &  & 227   &  & 5 280 707.35  & -33.15 &  & 5 560 969.73  & -31.37 &  & 44 548     \\
       & Struct-CKN\textbf{+}       &  & \textbf{4 331 589}  & \textbf{-46.49} &  & \textbf{1803} &  & \textbf{89}    &  & \textbf{4 323 521.96}  & \textbf{-45.27} &  & \textbf{4 386 135.29}  & \textbf{-45.87} &  & \textbf{25 196}     \\
       &          &  &          &        &  &         &  &      &  &             &        &  &             &        &  &            \\
       & GENCOL-DCA &  & 6 778 925  &        &  & 2513 &  & 187   &  & 6 610 562.32  &        &  & 6 939 096.41  &        &  & 29 368     \\
       & CNN &  & 4 763 520  & -29.73 &  & 4940 &  & 318   &  & 4 697 990.54  & -28.93 &  & 4 947 363.87  & -28.70 &  & 55 263     \\
6      & Struct-CKN       &  & 5 074 211  & -25.15 &  & 3222 &  & 278   &  & 5 003 421.94  & -24.31 &  & 5 368 766.13  & -22.63 &  & 51 166     \\
       & Struct-CKN\textbf{+}       &  & \textbf{4 356 910}  & \textbf{-35.73} &  & \textbf{1961} &  & \textbf{126}   &  & \textbf{4 349 221.21}  & \textbf{-34.21} &  & \textbf{4 491 319.98}  & \textbf{-35.28} &  & \textbf{26 529}     \\
       &          &  &          &        &  &         &  &      &  &             &        &  &             &        &  &            \\
\hline
       &          &  &          &        &  &         &  &      &  &             &        &  &             &        &  &            \\
       & GENCOL-DCA &  & 9 482 896  &        &  & 2640 &  & 184   &  & 9 332 218.23  &        &  & 9 649 205.46  &        &  & 27 551     \\
       & CNN &  & 8 120 656  & -14.37 &  & 4765 &  & 341   &  & 8 009 867.62  & -14.17 &  & 8 540 997.09  & -11.48 &  & 57 435     \\
Mean   & Struct-CKN       &  & 6 443 520  & -32.05 &  & 3126 &  & 238   &  & 6 346 324.34  & -32.00 &  & 6 663 977.62  & -30.94 &  & 38 691     \\
       & Struct-CKN\textbf{+}       &  & \textbf{4 143 236}  & \textbf{-56.31} &  & \textbf{2000} &  & \textbf{119}   &  & \textbf{4 131 341.64}  & \textbf{-55.73} &  & \textbf{4 220 414.97}  & \textbf{-56.26} &  & \textbf{24 863 }   \\
\hline
\end{tabular}}
\label{table_solver_results_appendix}
\end{table}

\begin{table}[H]
\centering
\caption{Computational Results for Monthly Solutions}
\resizebox{\textwidth}{!}{
\begin{tabular}{cC{0.1mm}rrC{0.1mm}rrC{0.1mm}r}
\hline
 Initial solution && Solution cost & Diff. vs GENCOL-DCA && Cost of & Diff. vs GENCOL-DCA && Number of deadheads \\
 and clusters &&           & (\%) &&     global constraints       & (\%) &&                    \\
 \cline{1-1} \cline{3-4} \cline{6-7} \cline{9-9}
GENCOL init &  & 30 681 120.5              &       &  & 9 465 982.28         &        &  &  1725               \\
GENCOL-DCA     &  & 20 639 814.6  &  -32.73 \scriptsize{(vs GENCOL init)} &  & 2 127 086.77 & -77.53 \scriptsize{(vs GENCOL init)}    &  & 992                 \\
\midrule
CNN &  & 18 881 977.9              & -8.52  &  & 465 687.94          & -78.11 &  & 1014                \\
Struct-CKN       &  & \textbf{18 677 641.1}               & \textbf{-9.51}  &  & \textbf{420 095.87}          & \textbf{-80.25}  &  & \textbf{915}                 \\
\midrule
CNN\textbf{+}            &  & 18 617 109.4               &   -9.80     &  &  333 983.69         &  -84.30    & & 997 \\
Struct-CKN\textbf{+}       &  & \textbf{17 145 543.4}               & \textbf{-16.93} &  & \textbf{58 797.21}           & \textbf{-97.24} &  & \textbf{583}                 \\
\hline
\end{tabular}}
\label{table_final_results_appendix}
\end{table}


\end{document}